\documentclass[]{interact}

\usepackage{epstopdf}% To incorporate .eps illustrations using PDFLaTeX, etc.
\usepackage[caption=false]{subfig}% Support for small, `sub' figures and tables

\usepackage{array}
\newcolumntype{C}[1]{>{\centering\let\newline\\\arraybackslash\hspace{0pt}}m{#1}}
\newcolumntype{L}[1]{>{\raggedleft\let\newline\\\arraybackslash\hspace{0pt}}m{#1}}
\newcolumntype{R}[1]{>{\raggedright\let\newline\\\arraybackslash\hspace{0pt}}m{#1}}
\usepackage{multicol}
\usepackage{multirow}
\usepackage{bigstrut}
\usepackage{threeparttable}
\usepackage{mathtools}
\DeclarePairedDelimiter\ceil{\lceil}{\rceil}

\usepackage[ruled,vlined]{algorithm2e}
\usepackage{verbatim}
\usepackage{comment}
\usepackage{color}
\usepackage[normalem]{ulem}
\usepackage{hyperref}

\usepackage[natbibapa,nodoi]{apacite}% Citation support using apacite.sty. Commands using natbib.sty MUST be deactivated first!
\theoremstyle{plain}% Theorem-like structures provided by amsthm.sty
\newtheorem{theorem}{Theorem}[section]

\theoremstyle{definition}

\theoremstyle{remark}

\begin{document}

%\articletype{ARTICLE TEMPLATE}% Specify the article type or omit as appropriate

\title{Discrete Simulation Optimization for Tuning Machine Learning Method Hyperparameters}

\author{
\name{Varun Ramamohan\textsuperscript{a}\thanks{CONTACT Varun Ramamohan. Email: varunr@mech.iitd.ac.in}, Shobhit Singhal\textsuperscript{a}, Aditya Raj Gupta\textsuperscript{a} and Nomesh Bhojkumar Bolia\textsuperscript{a}}
\affil{\textsuperscript{a}Department of Mechanical Engineering, Indian Institute of Technology Delhi, Hauz Khas, New Delhi 110016, India}
}

\maketitle

\begin{abstract}
An important aspect of implementing machine learning (ML) methods involves controlling the learning process for the ML method so as to maximize the performance of the method. Hyperparameter tuning involves selecting a suitable set of ML method parameters that control its learning process. Given that hyperparameter tuning has been treated as a black box optimization problem subject to stochasticity, simulation optimization (SO) methods appear well suited to this purpose. Therefore, we conceptualize the hyperparameter tuning process as a discrete SO problem and demonstrate the use of the Kim and Nelson (KN) ranking and selection method, and the stochastic ruler (SR) and the adaptive hyperbox (AH) random search methods for solving said problem. We also construct the theoretical basis for applying the KN method, which determines the optimal solution with a statistical guarantee via solution space enumeration. The random search methods typically incur smaller computational overheads and can be proved to asymptotically converge to either global optima (the SR method) or to local optima (the AH method). We demonstrate the application of the KN and the SR methods to a wide variety of machine learning models, including deep neural network models. We benchmark the KN, SR and the AH methods with multiple state-of-the-art hyperparameter optimization (HPO) methods implemented via widely used libraries such as $hyperopt$, $mango$, and $optuna$. The performance of the SO methods indicates that they can provide highly competitive alternatives to existing HPO methods.%All three simulation optimization methods consistently outperform $hyperopt$'s random search method, and offer statistically comparable performance with respect to $hyperopt$'s Tree of Parzen Estimators method, the $mango$ reinforcement learning based method and the $optuna$ genetic algorithm.%used for time series prediction and image classification %The \textcolor{red}{SR and AHA methods outperform the $hyperopt$ RS method and offers statistically comparable performance with respect to $hyperopt$'s TPE method, the $mango$ reinforcement learning based method and the $optuna$ genetic algorithm.}%Machine learning (ML) methods are used in most technical areas such as image recognition, product recommendation, financial analysis, medical diagnosis, and predictive maintenance. 
\end{abstract}


\section{Introduction \& Literature Review}
\label{sec:intro}

\textit{Study Overview.}\\

Most machine learning (ML) algorithms or methods are characterized by multiple parameters that can be selected by the analyst prior to starting the training process and are used to determine the ML model architecture and control its training process \citep{kuhn2013applied,jordan2015machine}. Such parameters are referred to as `hyperparameters' of the ML method, in contrast to parameters of the ML method or model itself that are estimated during the training process, such as the slope and the intercept of a linear regression model estimated via a maximum likelihood estimation process applied on the training dataset for the problem. Examples of ML method hyperparameters include the support vector machine (SVM) classifier in the Scikit-learn Python ML library \citep{scikit-learn} that is controlled via hyperparameters such as the kernel type, regularization and kernel coefficient parameters, among others. Similarly, a feed-forward artificial neural network (ANN) method is controlled by the learning rate, learning type, optimization solver used (e.g., stochastic gradient descent, ADAM), the number of layers, and the number of nodes in each layer. It is evident that finding the optimal set of hyperparameters - that is, the hyperparameter set that maximizes a measure of the performance of the ML method appropriate to the prediction problem at hand - thus becomes important. Brute force enumeration of the solution space may not be a computationally tractable approach towards determining the optimal hyperparameter set, even if hyperparameters with continuous search spaces are discretized. Hence many approaches for hyperparameter optimization (HPO) or `tuning' have been proposed, such as grid searches, manual searches, metaheuristics, and Bayesian optimization \citep{yang2020hyperparameter}. In this paper, we conceptualize the process of hyperparameter tuning as a discrete simulation optimization (SO) problem and demonstrate the use of methods commonly used to solve discrete SO problems for HPO.

SO methods are used to solve optimization problems wherein the objective function values are estimated via a simulation \citep{henderson2006handbooks}. They can be expressed as follows. 

\begin{equation}
	\label{eqsimopt}
	\centering
	\underset{x \in \mathbb{S}}{max}~E[h(x, Y_x)]
\end{equation}

In the above equation, $h(x, Y_x)$ represents the simulation outcome from one replication of the simulation, and we are interested in maximizing its expected value. $x \in \mathbb{R}^n$ represents the vector of decision variables taking values in the feasible search space $\mathbb{S}$ ($\mathbb{S} \subseteq \mathbb{R}^n$), and the $Y_x$ are random variables representing the stochasticity in the problem. In this paper, we conceptualize the training and validation process of the machine learning method as the simulation, and its hyperparameters as the decision variables. The outcome of a single replication of the `simulation' is represented by a performance measure - such as the classification accuracy or the area under the receiver operating characteristic curve - of the machine learning method on a validation dataset, and thus the objective function becomes the average value of the performance measure that we wish to maximize. In Section~\ref{theory}, we describe how we model stochasticity in the ML method training and validation process. This study is a first step towards demonstrating that SO methods can be leveraged to perform HPO. A key advantage of many SO methods, over commonly used hyperparameter tuning methods such as grid search, random search, etc., is that they explicitly consider stochasticity in the optimization problem, and provide statistical guarantees of finding the optimal solution (i.e., the hyperparameter set) via enumeration within finite computational budgets, or provably possess the property of asymptotic convergence to the  global or local optimal solutions. This implies that if sufficient computational overheads are available for the ML method training process, then SO methods are highly likely to find hyperparameter sets that are at least locally optimal.

As a first step towards leveraging SO methods for this purpose, we focus on HPO problems with discrete search spaces, and demonstrate the use of two types of SO methods for our conceptualization of HPO as a discrete SO problem: ranking and selection (R\&S) methods and random search methods. R\&S methods are developed for discrete search spaces, and random search methods may be used for both discrete and continuous search spaces. We present the numerical illustration of our approach in two phases. In the first phase, we illustrate the use of the Kim and Nelson (KN) R\&S method \citep{kim2001fully}, and the stochastic ruler (SR) random search method \citep{yan1992stochastic} for HPO in a wide variety of ML settings. For instance, we consider the hyperparameters of SVMs and feedforward ANNs for a classification problem with a standard dataset, and also consider HPO for long short term memory (LSTM) and convolutional neural networks (CNNs) for time series and image datasets, respectively. In the second phase, we benchmark the performance of the KN and the SR methods, and also a locally convergent random search (LCRS) method developed for higher-dimensional problems, against multiple existing methods representing the major types of HPO approaches. We choose to benchmark the SO methods against mature implementations of these HPO methods within widely used state-of-the-art hyperparameter tuning packages such as $hyperopt$, $mango$, and $optuna$. We also address a theoretical issue associated with applying many R\&S methods to the HPO problem, which requires that the simulation outcomes from each replication possess the $iid$ property.

We now discuss the relevant literature and our associated research contributions.\\ %\textcolor{red}{As part of this, we also discuss the rationale underlying our choices of SO techniques and baseline methods for the benchmarking experiments.}\\

\noindent \textit{Literature Review.}\\

Multiple HPO methods, and indeed multiple different types of HPO methods, have been developed. The majority of HPO methods have been developed for single-objective HPO, wherein the objective function typically involves maximizing a single-valued function of ML method performance \citep{bischl2021hyperparameter,yang2020hyperparameter,yu2020hyper}. Further, in the very recent past, HPO has been conceptualized as a multi-objective problem as well \citep{karl2022multi}. As an example, \cite{karl2022multi} argue that multiple types of misclassification costs must be taken into account in the context of ML applications in medical diagnostics, making it challenging to capture all such costs in a single objective. However, given that this study represents a first step towards the consideration of SO methods for HPO, we consider only methods developed for the classical single-objective HPO problem, while acknowledging that SO methods exist for multi-objective problems as well \citep{cooper2020pymoso}.%In this section, we provide an overview of the different extant HPO approaches.

Broadly, existing HPO methods can be considered to be of the following types: simple model-free enumerative or sampling-based methods such as grid search or random search; sequential Bayesian optimization methods; gradient-based methods; multi-fidelity methods; and metaheuristics \citep{yang2020hyperparameter,bischl2021hyperparameter}. We discuss each of these HPO method classes below.

We begin by consider the hyperparameter tuning method known as `babysitting', which is essentially a trial and error method. This method is used by analysts with prior experience with the prediction problem at hand, and is useful when severe limitations on computational infrastructure are present \citep{yang2020hyperparameter}. However, this approach cannot be generalized across ML methods, especially when the number of hyperparameters are relatively large, and rarely identifies an optimal or near-optimal solution.

The grid search (GS) method is one of the first and most popular systematic procedures used for HPO \citep{bergstra2011algorithms}. The method involves discretizing the hyperparameter space, and then exhaustively evaluating the average performance of each solution in the discretized hyperparameter space via, for example, cross validation. However, the method does not provide a statistical guarantee of selecting the optimal hyperparameter set within a finite computational budget. 

To address the limitations of the GS method, \citet{bergstra2012random} proposed the `random search' (distinct from SO random search methods) HPO approach, and showed that it outperformed GS for the same computational budget. The method involves defining the hyperparameter space as a bounded set - typically discrete - and then samples hyperparameter sets randomly from said space. The methodology does not consider information from prior iterations, and thus poorer performing regions of the space also receive equal consideration in the search. Similar to GS, the approach does not consider stochasticity in the HPO problem explicitly and also does not provide statistical optimal solution selection guarantees. %hough its time complexity is $O(n)$ (where $n$ is the size of the grid), the search methodology is still independent of the prior observations and thus wastes resources evaluating even the poorer performing regions equally. Moreover, it can be only used on a discrete search space.

Another popular technique for optimizing hyperparameter selection is the Bayesian Optimization (BO) method. BO involves performing relatively fewer function evaluations as it uses information from previous iterations to select the new solution at each evaluation it uses previously obtained observations to determine next evaluation points. \cite{snoek2012practical} introduced BO for HPO and demonstrated its application for several ML algorithms. It uses an acquisition function, typically a Gaussian process, to model the objective function and determines the next candidate for the optimal solution by predicting the best system according to the acquisition function. The acquisition function is updated at each iteration based on information from previous objective function evaluations. Some drawbacks of BO include sensitivity to the choice of acquisition function and its parameters and cubic time complexity with number of hyperparameters. Moreover, it has difficulty handling categorical and integer-valued hyperparameters. Examples of BO based methods include the random forest based and Tree of Parzen Estimators based HPO methods \citep{bergstra2013making,eggensperger2015efficient}.
%\cite{gardner2014bayesian} used BO with inequality constraints called constrained Bayesian Optimization (cBO) for optimizing hyperparameters.

Stochastic gradient based techniques have been used to explore continuous hyperparameter spaces. For example, \citet{maclaurin2015gradientbased} and \citet{pedregosa2016hyperparameter} demonstrated the use of the stochastic gradient descent method for finding optimal values of continuous hyperparameters of various machine learning algorithms.

Multi-fidelity methods have been developed to address the issue of limited computational resources, taking cognizance of the fact that the HPO process can be computationally expensive depending on dataset size and hyperparameter space \citep{yang2020hyperparameter}. Such methods expend different amounts of computational effort on poor-performing and promising regions of the search space, balancing the exploration versus exploitation trade-off. Most examples of such methods applied to HPO problems are bandit-based methods, such as successive halving \citep{karnin2013almost}, and \textit{hyperband} \citep{li2017hyperband,falkner2018bohb}. \cite{sandha2020mango} develop a state-of-the-art bandit-based approach, which consists of an efficient and effective batch Gaussian process bandit search approach using upper confidence bound as the acquisition function.

Another emerging class of approaches involves applying reinforcement learning (RL) methods to HPO \citep{jaafra2019}. \citet{jomaa2019hyp} conceptualized the HPO problem as a sequential Markov decision process with expected loss on the validation dataset as the value function. The use of model-based and context-based meta-RL methods for HPO have been explored in a series of recent studies \citep{wu2020eff,liu2022con,wu2023hyp}, including in a multi-objective framework that incorporates considerations around CPU utilization and latency as well \citep{chen2021emorl}. \citet{dong2019vis} developed a custom deep continuous Q-learning based RL approach for optimizing the hyperparameters associated with automated visual object trackers. Similar RL-based custom approaches have been developed for deep learning applications in cryptography \citep{rijsdijk2021RL} and semantic segmentation \citep{iranfar2021RL}. The RL approaches towards HPO are diverse in nature, in that they may be model-free, model-based or context-aware, and hence a single dominant RL approach towards HPO has not emerged that has been incorporated into commonly used HPO libraries. Further, the above RL approaches do not explicitly consider stochasticity in the ML method accuracy measurement process for a given hyperparameter set, despite the fundamentally stochastic nature of the RL framework. For example, while \citet{jomaa2019hyp} conceptualize the hyperparameter set selection process within their RL framework as a Markov decision process, the transition probabilities for the Markov process do not, in our understanding, consider the uncertainty in the ML method performance measure estimate generated via the training and validation process associated with a given hyperparameter set.

Metaheuristic algorithms have been used for HPO as well. Among these, the genetic algorithm (GA) and particle swarm optimization (PSO) methods have been used most widely. For example, \citet{lessmann2005optimizing}, \citet{hutter2011sequential} and \citet{peng2004model} applied a GA to optimize support vector machine hyperparameters, and \citet{lorenzo2017pso} applied the PSO approach to optimize hyperparameters of deep neural networks. These approaches have drawbacks similar to that of other HPO approaches: they do not explicitly account for problem stochasticity, and do not provide statistical guarantees of optimal solution selection.

Multiple Python libraries for HPO exist. Widely used among these include the $hyperopt$ package \citep{bergstra2013making}, which implements the Tree of Parzen Estimators (TPE) based BO approach. It also implements discrete and continuous versions of random search (RS) HPO methods \citep{bergstra2012random,bergstra2013making}. The $hyperopt$ is integrated within the popular ML frameworks $scikit-learn$ \citep{scikit-learn} as well as $keras$ Tensorflow \citep{chollet2015keras}. The bandit-based algorithm by \cite{sandha2020mango} is available as the Python library $mango$. The $keras-tuner$ package, which is developed for the Tensorflow ML framework, is another popular package for deep neural network learning practitioners. Finally, one of the most popular packages is $optuna$, which provides implementations of the grid search and random sampling search methods, Bayesian optimization methods, and a genetic algorithm metaheuristic, among others \citep{optuna_2019}. We now discuss key challenges associated with HPO, how SO methods can help address these challenges, and our research contributions within this context.\\%As will be discussed in detail in Section~\ref{orgcomp}, we choose implementations of the different types of the previously discussed HPO methods within these packages for benchmarking the SO algorithms that we implement for HPO.%Another commonly used state-of-the-art library is $mango$, which implements an efficient and effective batch Gaussian process bandit search approach using upper confidence bound as the acquisition function \citep{sandha2020mango}.

\noindent \textit{Research Contributions.}\\

HPO has often been described as a black-box optimization problem, implying that many of the methods developed for optimization problems that involve knowledge of the problem structure may not be applicable. SO methods are thus likely to be well suited for HPO problems. More importantly, as discussed in multiple reviews \citep{yang2020hyperparameter,bischl2021hyperparameter}, a key challenge associated with HPO problems involves their inherent stochasticity, given that ML method accuracies (for a given hyperparameter set) may vary depending on multiple factors such as dataset organization, training algorithm initialization, etc. SO methods, given that they are by definition developed to find optimal solutions while accounting for stochasticity in the optimization problem, are thus well suited to address this challenge associated with HPO. Finally, ML method performance evaluations can be computationally expensive depending upon the size of the dataset \citep{yang2020hyperparameter,yu2020hyper}. Many SO methods are developed to account for expensive objective function evaluations. For example, these include optimal computing budget allocation methods \citep{chen2011stochastic}, or methods that attempt to maximize computing time spent on promising regions of the search space \citep{hong2006discrete,xu2013adaptive}. Thus, SO methods are once again well suited to address this challenge associated with HPO.

In light of the above challenges associated with HPO, and the potential of SO methods to address these challenges, the key research contribution of this study is the conceptualization of the ML training and holdout validation process as a simulation, and further, that of HPO as a SO problem. While SO methods have been developed to work in both discrete and continuous search spaces, we chose to consider only discrete search spaces for this paper as a first step. Thus, another research contribution involves the demonstration of methods commonly used for discrete SO problems for HPO. As part of this, we show how R\&S methods such as the KN method can be applied to the HPO problem and construct the theoretical grounding for its application. We then show how the globally convergent (in probability) SR random search method \citep{yan1992stochastic} and one of its novel variants \citep{ramsr2020} can be applied to the HPO problem, including the construction of two different neighborhood structures and stochastic rulers. Finally, as part of the benchmarking exercise, we demonstrate the use of the adaptive hyperbox (AH) LCRS method \citep{xu2013adaptive}, developed for relatively large search spaces (i.e., exceeding 15-20 dimensions), for an HPO problem with a relatively large search space. From the SO standpoint, our demonstration of how the KN method can be applied in situations with a computational budget specified in terms of maximum number of simulation replications is, in our knowledge, novel. Further, the large variety of HPO problems in different ML settings promises to serve as a rich test bed for SO methods, and can motivate development of newer methods, as we discuss in Section~\ref{conc}.

%In our search of the literature, we have not identified another study that utilizes discrete simulation optimization methods for ML method hyperparameter optimization. As mentioned earlier, the key advantage of applying R\& S methods over methods such as GS and random search is that they provide a probabilistic guarantee of selecting the best hyperparameter set - for example, they guarantee that the optimal solution will be selected with 1 - $\alpha$ probability as long as the difference between its objective function value and that of the next best solution is at least $\delta$. Similarly, random search methods such as the stochastic ruler method provide asymptotic guarantees of converging to the optimal solution. We provide a conceptualization of the ML training and holdout validation process as a simulation, and also construct the theoretical grounding for applying the KN ranking \& selection procedure for hyperparameter optimization. Finally, we provide a detailed benchmarking of the KN R\&S method and the stochastic ruler random search method against popular hyperparameter optimization methods implemented in the $hyperopt$ and $mango$ libraries.

The remainder of the paper is organized as follows. In Section~\ref{theory}, we set up the theoretical framework for applying SO methods to HPO. In Section~\ref{methodology}, we provide an overview of the computational experiments along with the rationale for our choice of SO methods and baseline HPO methods for the benchmarking exercise (Section~\ref{orgcomp}), as well as an overview of the SO methods we implement (Sections~\ref{knover} and \ref{srover}). In Section~\ref{compres}, we describe the computational experimentation itself, and we conclude the study in Section~\ref{conc} with a discussion of the potential impact and limitations of our work and avenues of future research.

%~\ref{kneval} and \ref{srnumres}, we describe the results of applying the KN R\&S method and the stochastic ruler random search, respectively, to optimize the hyperparameters of various ML methods. We also provide benchmarking results for these methods against $hyperopt$ and $mango$ hyperparameter optimization methods. We conclude the paper in Section~\ref{conc} with a discussion of the potential impact of our work, its limitations, and avenues for future research.

\section{Theoretical Framework}
\label{theory}
Consider a machine learning problem where we are attempting to find the best-fit function \(f\) that predicts an outcome \(y\) as a function of features \(x\). Here \(x ~  \epsilon ~  \mathbb{R}^n\), \(y ~ \epsilon ~  \mathbb{R}\). Let the training dataset be denoted by \((X,Y)\), where \(X\) is an \(m \times n\) matrix and \(Y\) is an \(m\)-dimensional vector of labels. Note that we do not make a distinction between classification and regression at this juncture, and hence we let $y \in \mathbb{R}$, and do not impose any other constraint on $y$. For example, if our focus was only on classification problems, we would specify that $y \in \{0,1,..,k-1\}$, where $k$ represents the number of classes. 

Let \(\theta ~ \epsilon ~ R^d\) be a set of \(d\) hyperparameters that characterize the machine learning model architecture and training process, and let \(P\) be the performance measure used to judge the quality of the fit of \(f(x)\) (without loss of generality, we assume increasing \(P\) denotes increasing quality of fit). Typically, \(P\) is a random variable whose value depends upon the organization of the training and holdout (test) validation datasets, and hence hyperparameter optimization attempts to find \(\theta\) that maximizes \(E[P]\). \(E[P]\) may be estimated, for example, via a cross-validation exercise. We propose to use discrete SO methods for selecting optimal hyperparameter values. More specifically, we demonstrate here the use of ranking and selection methods such as the KN method \citep{kim2001fully} and random search methods such as the stochastic ruler method \citep{yan1992stochastic} and one of its variants \citep{ramsr2020} for this purpose. 

\subsection{Simulation Algorithm}
\label{theorysim}
In order to apply SO methods to find optimal hyperparameter values for ML methods, we first need to define the corresponding `simulation' and the associated optimization problem. We propose considering the process of finding the optimal parameter estimates of the function \(f\) given a particular organization of the training set \((X, Y )_i\) and a specific \(k_{th}\) set of hyperparameters \(\theta_k\), and evaluating the quality of the fitted function $f$ via holdout validation by generating an estimate of the performance measure, as the \(i_{th}\) replication of the simulation. The particular version of the training set used in the \(i_{th}\) replication is generated by a single permutation of the dataset, denoted by \(\sigma_i (X, Y )\), where $\sigma$ is the permutation function that operates on the dataset. The permuted dataset \((X, Y )_i\) can then be divided into the training and holdout validation datasets (e.g., first 80\(\%\) used for training, next 20\(\%\) for holdout validation). Thus the output of one replication (e.g., the \(i_{th}\) replication) of this simulation can be considered to be the performance measure \(P_i (\theta_k)\), denoted in short as \(P_{ki}\). The simulation algorithm can thus be written as follows, depicted in Algorithm~\ref{A1}. In Algorithm~\ref{A1} below, in order to make the simulation conceptualization clear, we also explicitly describe the permutation process (Step 3).%Each permutation can be considered to be a random mapping - i.e., if the set of indices of the \(m\) samples in the training set is denoted by \(M =\) \{1, 2,..., m\}, then the permutation function \(\sigma\) randomly samples (without replacement) an index from the training set and maps it to the set \(M\).\\

\begin{algorithm}[H]
	\caption{Simulation algorithm for generating an estimate of \(E [P]\) for a given hyperparameter set \(\theta_k\).}
	\label{A1}
	\SetAlgoLined
	%\KwResult{Write here the result }
	1. Initialize with dataset $(X, Y )$, number of replications $I$ used to estimate $E[P]$ for a given hyperparameter set $\theta_k$, set of indices of the dataset samples \(M =\) \{1, 2,..., m\}, hyperparameter set $\theta_k$, $P( \theta_k ) = 0$. \footnotesize /* Comment. For example, if the dataset size is 100, then $M = \{1,2,\dots,100\}$, with $m = 100$.*/\\\normalsize
	2. Set random number seed $s_k \sim g_s(s)$. ~~~~~\footnotesize /* Comment. Sample random number seed from its \(cdf\) $g_s(s)$ for generation of \(\overline{P(\theta_k)}\) */\\ \normalsize
	3. Permute dataset and estimate $E[P]$ (as $\overline{P(\theta_k}$) given hyperparameter set $\theta_k$.\\
	\For{\(i = 1-I\)}{
		\((X, Y )_i = \{\}\), \(J = |M|\)\\
		\For{\(j = 1-J\)}{
			\(m \leftarrow |M|\)\\
			\( u \leftarrow \text{Sample from } U(0,1) \) \footnotesize /* Sample $u$ from uniform random variable on (0,1). */ \\ \normalsize
			\(l \leftarrow \lceil u \times m \rceil \) ~~~~~ \footnotesize /* Randomly sample $l^{th}$ element from $M$. */  \\
			\normalsize \((X, Y )_i \leftarrow (X, Y )_i \cup (x_l, y_l )\) \footnotesize /* $(x_l, y_l)$ is the $l_{th}$ sample from dataset $(X, Y)$. \normalsize \\
			%\(M \leftarrow M- \{l\}\)\\
			$m \leftarrow |M - \{l\}|$\\
			\(M \leftarrow \{1,2,\dots,m\}\)
		}{
			\(P(\theta_k) = P(\theta_k) + fit((X, Y )_i, f, \theta_k)\)\
		}
	}{
		\(\overline{P(\theta_k)} \leftarrow \frac{P(\theta_k)}{I}\)}  \\
	
\end{algorithm}

In Algorithm \ref{A1}, the term \(fit((X, Y )_i, f)\) can be thought as the training and validation subroutine that takes as input the \(i_{th}\) permutation of the dataset \((X, Y )\), the function to be fit \(f\) and the hyperparameters \(\theta_k\). The subroutine then divides this dataset into training and validation subdatasets, estimates the parameters of $f$ using the training subdataset, and then outputs the performance measure \(P_{ki}\) for the dataset \((X, Y )_i\) by applying the fitted function on the validation dataset. We note here that while the generation of the training and validation subdatasets itself from a given dataset (say, dataset $(X,Y)_i$) itself can involve some stochasticity, we assume otherwise - for example, we assume that the first, say, X\% of the dataset is always used as the training subdataset and the remaining 100-X\% of the dataset is then used as the validation subdataset. Thus the generation of the training and validation subdatasets is also determined by the random number seed $s_k$ specified in Algorithm~\ref{A1}.
%\section{De Finetti's theorem}
%To establish the IID property of the sequence of performance measure generated by training the model on different random permutations and combinations of the dataset. The arrangement of the training and validation set depends on the random number seed

%\subsection{Optimization Problem}
Now that we have conceptualized the generation of the $i^{th}$ estimate of the performance measure $P_{ki}$ as one replication of the simulation, we can write down the optimization problem as follows.

\begin{equation}
	\label{eqmlsim}
	\underset{\theta \in S}{max}~ E[P(\theta)]   
\end{equation}

Here \(S\) represents the set of allowable values for \(\theta\). \(E[P(\theta)]\) is estimated as \(\overline{P(\theta)}\) by the simulation described above.

We now discuss how the set of $I$ replicate values of $P_{ki}$ generated by Algorithm~\ref{A1} used to estimate $E[P(\theta_k)]$ can be considered to be an $iid$ sample despite the fact that they are generated from effectively the same dataset $(X,Y)$.

\subsection{\(IID\) Requirements}
In order to apply common R\&S methods such as the KN or NSGS procedures \citep{henderson2006handbooks}, or random search methods such as the stochastic ruler algorithm \citep{yan1992stochastic}, a standard \(iid\) requirement must be satisfied by the simulation replications associated with a given hyperparameter set \(\theta_k\). This assumption is formalized below.

\noindent \textit{Assumption 1.} The sample $P(\theta_{k1}), P(\theta_{k2}), \dots, P(\theta_{kI})$ used to estimate $E[P(\theta_k)]$ for the $k_{th}$ hyperparameter set $\theta_k$ $(k \in \{1,2,\dots,K\})$ via Algorithm~\ref{A1} is \(iid\).%That is, the \(I\) replications \(P_{k1}, P_{k2},..., P_{kI}\) must be \(iid\).

Note that, if \(|S| = K\) (that is, there are \(K\) allowable values of \(\theta\)), some procedures such as the NSGS method impose the additional requirement that the random variables \(\overline{P(\theta_k)}\), \(k = (1, 2,..., K)\) are also independent; however, other methods such as the KN method do not, as they permit the use of common random numbers \citep{kim2001fully,henderson2006handbooks}. Hence our focus in this section will be
the \(iid\) requirement that the \(P_{ki}\), \((i = 1-I)\) have to satisfy.

The above assumption is required for the KN method due to the following reason. Assume that the $K$ solutions $\theta_k, ~ k \in \{1,2,\dots, K\}$ in the feasible region $S$ are indexed such that $\mu_K > \mu_{K-1} > \mu_{K-2} > \dots > \mu_{1}$. The KN method guarantees that, out of the $K$ solutions in the feasible region $S$, the best solution is selected with probability $1 - \alpha ~(0 < \alpha < 1)$ as long as the next best solution is at least $\delta ~(\delta > 0)$ less than the best solution. $\delta$ specifies the indifference zone; that is, the experimenter is assumed to be `indifferent' to solutions with mean performances within $\delta$ of the best solution. $\delta$ and $\alpha$ are parameters of the KN method and are specified prior to application of the method. The theoretical underpinning of the KN method - that is, the result that guarantees that the best solution $\theta_K$ is selected with probability $1 - \alpha$ given $\delta$ is provided below. As can be seen from its statement, it requires \textit{Assumption 1} above.

\begin{theorem}
		(Adapted from Theorem 1 from \cite{kim2001fully}.) Let $X_1, X_2,..$ be iid multivariate normal with unknown mean vector $\mu$ such that $\mu_K \geq \mu_{K-1} + \delta$ and unknown and positive definite covariance matrix $\Sigma$. Then with probability $1 - \alpha$, the KN method selects solution $\theta_K$.
\end{theorem}

In the above result, $X_j = X_{1j}, X_{2j}, \dots, X_{Kj}$ represents the $j_{th}$ vector of observations across all solutions $\theta_1, \theta_2, \dots, \theta_K$.

For typical machine learning exercises, the samples in a given organization of the dataset \((X, Y )_i\) are assumed to be \(iid\). However, this is typically difficult to verify, especially if the analyst is not involved in the generation of the dataset. In such situations, the operational definition of the \(iid\) property to be applied is derived from the class of representation theorems discussed in \citet{Ressel1985} and \citet{oneillexchangeability}. Broadly, according to these theorems, a sequence (typically infinite) of random variables (e.g., \(X_1, X_2,... \)) consisting of terms that may not be independent of each other can be shown to be conditionally independent given a realization of a random variable $Q$ driving the generation of the sequence as long as the terms in the sequence are exchangeable. The key requirement here is exchangeability - that is, given a realization of $Q$, the joint distribution of the sample must not change if the order in which the \(X_i\) are generated changes \citep{Ressel1985}.

For example, consider a sequence of random variables $X_1, X_2,\dots$ that are generated independently from the uniform distribution $U(a,b)$. Now if $a$ is a constant, and $b$ is itself a Bernoulli random variable that takes value $b_1$ with probability $p$ and $b_2$ with probability $1-p$, then the distribution of the sequence of random variables is driven by the realization of the random variable $b$. The representation theorems assert that the $\{X_i\}$ can be considered to be an $iid$ sample as long as the order in which the $X_i$ are generated does not change their joint distribution. Note that while the original representation theorem due to De Finetti was developed for infinite sequences of Bernoulli random variables, a version of the representation theorem for finite sequences was developed by \citet{diaconis1980}.

In order to demonstrate the applicability of this class of theorems to our case, we must identify the distribution that drives the generation of the particular set of replications \(P_{ki}, i\) = 1 \text{ to } \(I\) and then demonstrate that these replicate values of \(P_{ki}\) are exchangeable.

To this end, we first note that each value of \(P_{ki}\) is a function of the \(i_{th}\) permutation of the training set. Each permutation of the training set \((X, Y )_i\) can be considered to be governed by the sequence of uniform random numbers that are used to sample (without replacement) from the training set \((X, Y )\). For example, per Algorithm 1, the \(i_{th}\) permuted dataset \((X, Y )_i\) is generated by the sequence of random numbers \(\{U_i\} = U_{i1}, U_{i2}, ..., U_{iJ} \). However, the \(\{U_i\}\) are a subsequence of the stream of pseudorandom numbers generated for the entire simulation in algorithm 1, which can be thought of as a sequence in itself, given by \(\{U\} = U_{1}, U_{2}, ..., U_{I}\). This sequence is governed by a specific seed \(s_k\) that is supplied to the random number generator at the start of the simulation. That is, if we denote the seed used to generate \(U\) as \(s_k\), then \(U\) in effect becomes a function of \(s_k\). Now, if the \(s_k\) are themselves generated from some distribution \(g_s(s)\) - for example the \(s_k\) are randomly sampled (with replacement) integers from 1 to \(L\), where \(L\) is a very large number when compared to \(K\) to minimize the likelihood of sampling the same \(s_k\) twice or more - then the underlying distribution governing the generation of the \(P_{ki}\) becomes this uniform random integer distribution. Thus we have constructed the distribution driving the generation of the \(P_{ki}\).

Exchangeability is now easy to demonstrate, given that each \(P_{ki}\) is a function of the random number seed \(s_k\). Therefore, the joint distribution of the \(P_{ki}\) depends only on the distribution of the \(s_k\), and not on the order in which the \(P_{ki}\) are generated. This completes the argument.

%We now describe the numerical experiments involving implementation of the KN ranking and selection method to the hyperparameter optimization problem.

\section{Study Methodology}
\label{methodology}
%Numerical Evaluation: KN Ranking and Selection Method
In this section, we describe the methodology used in this study to demonstrate and evaluate the use of SO methods for solving the HPO problem. As part of this, in Section \ref{orgcomp}, we describe the overall approach followed for the computational experimentation and benchmarking process, and in Sections~\ref{knover} and \ref{srover}, we outline the KN and the SR random search methods. In order to contain the length of the manuscript, we provide a description of the AH LCRS algorithm in Appendix~\ref{ap:aha}.

\subsection{Overview of Computational Experimentation}
\label{orgcomp}
Our strategy for demonstrating the utility of SO methods for solving HPO is as follows. We first demonstrate the application of SO methods for solving a wide variety of HPO problems with discretized search spaces. We then benchmark the SO methods implemented in this study, based on a single classification task but with different hyperparameter search spaces and computational budget constraints, against widely used state-of-the-art HPO methods. For all benchmarking exercises presented in this study (i.e., for the KN, SR and AH methods), we consider a realistic hyperparameter tuning situation, where the optimal hyperparameter set is not known previously, and hence the optimal hyperparameter set is typically determined in conjunction with constraints such as computational budgets specified in terms of the number of function evaluations or computational runtime.

We now provide the rationale underlying our choice of SO methods and HPO methods used in this study.

We first note that only HPO problems with discretized search spaces are considered in this study and have chosen SO methods accordingly. We acknowledge that while such problems can have continuous or mixed-integer search spaces, they are beyond the scope of this paper given that this is a first step towards leveraging SO methods for HPO. Thus, in this paper, we chose to focus on: (a) providing a conceptualization of the ML training and validation process as a simulation; (b) addressing potential theoretical obstacles towards applying basic SO methods such as R\&S; and (c) providing a comprehensive demonstration and benchmarking of the SO methods developed for or typically used for discrete search spaces. We anticipate that once it is established that SO methods provide a competitive alternative to existing methods for solving discrete HPO problems, then it is likely that sufficient interest will be generated in the SO and potentially the ML HPO community as well to pursue the natural extension of this work to HPO problems with continuous or mixed-integer search spaces.

Given that this study focuses on discrete HPO problems, we chose the KN and the SR method for most of the computational experiments, with the AH method employed only in the benchmarking process. This is due to the following reasons. First, the KN R\&S method and the SR method represent two key paradigms in discrete SO: fully enumerative ranking and selection, and random search methods with asymptotic convergence to the global optimum. Second, given that this study represents the first step towards applying SO methods for HPO, we wished to show that if these relatively simple methods in each discrete SO paradigm perform comparably to the state-of-the-art HPO methods, then it is likely to stimulate interest in applying more advanced SO methods for the HPO problem. Third, given that only a few hyperparameters form the focus of many HPO exercises (an average of approximately four, as demonstrated in \cite{yang2020hyperparameter}), these relatively simple methods are likely to be sufficient to provide at minimum comparable performance to existing methods for small to moderate HPO search spaces. Additionally, for the demonstration stage of the computational experiments, these simple methods were chosen to maintain focus on the application of these methods to the HPO problem rather than on the methods themselves.

The AH LCRS algorithm was chosen only for the benchmarking process for two reasons. First, because one of the problems considered in the benchmarking process involved a larger (in terms of cardinality) as well as higher-dimensional search space, an SO method developed specifically to handle such cases was required. In this context, the AH method was found suitable given its superior performance for higher-dimensional search spaces in comparison even with methods such as COMPASS or industrial strength COMPASS \citep{hong2006discrete,xu2010industrial,xu2013adaptive}. Secondly, the AH method was chosen to represent the LCRS paradigm in the random search SO literature.
 
For the benchmarking process, we chose to evaluate the SO methods against at least one method from each major HPO method class (discussed in the literature review in Section 1). Further, we chose to compare against implementations of these methods in widely used libraries (Python libraries in particular) given that most practitioners were unlikely to use HPO methods published in the literature that were not associated with mature implementations via software libraries. Thus, we chose to compare with: (a) the RS method, as implemented in the $hyperopt$ library, representing simple model-free methods; (b) the TPE Bayesian optimization method implemented in the $hyperopt$ library, representing Bayesian optimization methods; (c) the $mango$ implementation of a bandit-based method representing multi-fidelity methods; and (d) the $optuna$ implementation of a genetic algorithm representing metaheuristic approaches for HPO.%This is similar to the practice of benchmarking simulation-based optimization methods against the $OptQuest$ SO package available on commercial simulation platforms \citep{xu2013adaptive,xu2010industrial,aboueljinane2022discrete}

The choice of HPO methods and their implementations in existing libraries for the benchmarking exercise was based on our prior experience, combined with information from a small survey that we conducted among ML practitioners. These included academic and industry professionals, undergraduate and postgraduate students. The survey received 42 valid responses. We first note that none of the respondents used methods based on published literature that did not have an associated mature implementation. A large proportion of respondents relied on babysitting (24\%) or brute force (21\%) searches without using any HPO libraries. Among those who used HPO libraries, $hyperopt$'s $scikit$-$learn$ implementation (\href{https://hyperopt.github.io/hyperopt-sklearn}{$hpsklearn$}) and its native implementation (24\%) and $optuna$ (19\%) were the most widely used. Interestingly, the majority of $hyperopt$ users among the respondents preferred using the grid search or the random search methods, and not the more sophisticated BO TPE method. The $keras-tuner$ package was also widely used (12\%) among those who preferred the Tensorflow ML framework. One respondent used a custom method, and the remainder (5\%) used software platform based libraries, such as those associated with the Matlab scientific computing platform. Note that some respondents used more than one of the above HPO libraries. Thus, we chose to work primarily with the $hyperopt$ and $optuna$ libraries. We did not work with the $keras-tuner$ library given its platform-specific nature. We used the RS and Bayesian TPE implementations of the $hyperopt$ library given their popularity of use, and $optuna$'s genetic algorithm was chosen as the metaheuristic benchmark given its relatively recent update. The $mango$ package was chosen because its bandit-based multi-fidelity method implementation given its relatively recent publication \citep{sandha2020mango} in comparison to others such as $hyperband$ \citep{falkner2018bohb}. We also note here that we chose not to use an RL method for the benchmarking process because, as discussed in in the literature review, none of the methods have found sufficient acceptance at the time of this writing to be included in widely used libraries such as $optuna$ or $hyperopt$. This can also be seen from the survey outcomes described above. Further, while the multi-fidelity bandit-based method $mango$ \citep{sandha2020mango} can primarily be considered as a Bayesian optimization HPO method, it has similarities with RL approaches in that it considers the exploration/exploitation trade-off in a dynamic manner as a function of computational resources expended.%Finally, the $optuna$ package was by far the most widely used package that we encountered in an informal survey of ML practitioners, and its genetic algorithm was chosen as the metaheuristic benchmark given its relatively recent update.

We now discuss the choice of ML tasks considered in this study. For the demonstration section of the computational experiments, we consider a wide variety of learning tasks so as to show the broad applicability of SO methods for HPO. These include binary classification tasks on numerical datasets using standard ML methods such as support vector machines and feedforward artificial neural networks, as well as `deep' learning settings such as time series prediction via long short term memory (LSTM) neural networks and image classification via convolutional neural networks. Note that the time-series prediction problem via LSTM neural networks represents the application of SO methods for HPO in regression settings which involve three-dimensional datasets - that is, each sample in the dataset is represented by a matrix wherein each row is a time-indexed sequence corresponding to each feature. Further, we consider multiple performance analysis scenarios for the above ML tasks in the demonstration section: scenarios where the optimal HPO set is known, and where it is unknown, in which case the mean improvement in ML method performance within a prespecified computational budget is considered. For the benchmarking task, we consider a standard binary classification task on a numerical dataset. We consider only a single classification task for the benchmarking process because our conceptualization of the ML training and validation process as a simulation is independent of the specific ML setting for which HPO is being performed. This can be seen from Section~\ref{theory}. Given that nearly all widely used HPO methods, including those considered for benchmarking our proposed HPO methods, employ such a black-box conceptualization of the HPO problem, it is likely that the nature (e.g., discrete, continuous, mixed-integer) and size of the solution space will be more relevant when comparing HPO methods than the ML setting itself. For example, if applying a deep convolutional neural network on a large image dataset is the ML setting under consideration, then the computational expense of a single function evaluation (i.e., generation of a single replicate value of the classifier performance measure) will be identical regardless of which HPO method is used. Hence, as described above, instead of considering multiple ML tasks for the benchmarking experiments, we consider multiple scenarios in terms of different HPO problem settings (problem size and computational budget constraints) associated with a single classification task.

We conclude this section by noting that in all ML settings considered, the number of hyperparameters that we consider does not exceed six. This is consistent with ML practice, wherein analysts are unlikely to focus on more than a few key hyperparameters well known to affect ML method performance. This is emphasized in the HPO method review by \cite{yang2020hyperparameter}, where the authors also considered an average of approximately four total hyperparameters in their computational experiments. Further, an average of less than three were considered `key' hyperparameters.

%Before we do so, we would like to note that the SR method is one of the simplest and one of the first developed random search simulation optimization methods, and hence our use of this method for hyperparameter optimization. Many other random search methods that can be used in a similar manner have been developed, such as other modifications to the SR method itself \citep{alrefaei2001modification,alrefaei2005discrete}, COMPASS \citep{hong2006discrete}, etc.

%In this section, we demonstrate the application of the KN R\&S method for ML hyperparameter optimization. We begin by providing an overview of the KN procedure, then demonstrate the application of the KN method to optimize the hyperparameters of various ML methods, and finally describe the results of benchmarking the KN method against widely used hyperparameter optimization methods implemented in the $hyperopt$ hyperparameter tuning package.

\subsection{KN Method Overview}
\label{knover}
The KN method is a partially sequential R\&S method wherein noncompetitive solutions are screened out in the first stage using an initial set of replications from each solution, and the remaining solutions are then evaluated and eliminated iteratively by generating one replicate observation of the objective function at each subsequent iteration. 

The KN method provides a guarantee that, out of $N$ possible hyperparameter sets (or feasible solutions) $\theta_n~ (n \in \{1,2,\dots,N\})$, the optimal set will be selected with a probability $1-p$, given that the difference between the best solution $\theta^*$ and the next best $\theta_n$ is at least $\delta$. The values of $p$ and $\delta$ are parameters of the algorithm. As part of the KN method, we generate an initial set of replications $R_0$ for each $\theta_n$, and then using $\delta$ and $p$, select a subset $D$ of the original $N$ feasible solutions. This is referred to as the screening phase. Only the variances of the pairwise differences between the replications of each hyperparameter set are used in determining the subset $D$ using the first set of $R_0$ replications. 

%In the KN method of Ranking and Selection class \citep{henderson2006handbooks}, we define the criterion for comparison which is discussed in the procedure below. If the outcome can not be clearly rejected, then we keep those values in the Indifference Zone(IZ). We keep on generating the results, till the number of systems left in the IZ is 1, that is the single best system. %Most of the hyperparameters for Machine Learning algorithms are discrete valued, so for hyper-parameter optimization, we started stochastic simulation with discrete optimization methods. Ranking and Selection is a class of discrete optimization methods, which keep comparing the outcomes at each stage and eliminates the outcome if it is clearly not the contender for the best performer. 

Once $D$ is identified after the screening phase, we generate an additional replication for each $\theta_n$, and the subset $D$ is updated using the screening criteria until $D$ contains a single hyperparameter set. This is the optimal $\theta_*$. This phase is referred to as ranking and selection. We provide the details of the procedure below, adapted from \citet{henderson2006handbooks}. 
%\begin{comment}

Step 1. Initialize the method with $R_0$, $\delta$ and $p$. Calculate $\eta$ and $h^2$ as:

\begin{equation*}
	\centering
	\begin{aligned}
		&\eta = \frac{1}{2}\left[\left(\frac{2p}{N - 1}\right)^{(2/(R_0 - 1))} - 1 \right], \; \text{and}\\ 
		&h^2 = 2\eta(R_0 - 1)
	\end{aligned}
\end{equation*}

Step 2. Generate $R_0$ replicate values of $P(\theta_n)$ for each $\theta_n$. Estimate their means $\overline{P(\theta_n)}$ for $n \in \{1,2,..,N\}$.

Step 3. Estimate the standard deviations of paired differences as follows:
\begin{equation*}
	S_{nl}^2 = \frac{1}{R_0 - 1} \sum_{j=1}^{R_0} (P(\theta_{nj}) - P(\theta_{lj}) - (\overline{P(\theta_n)} - \overline{P(\theta_l)}))^2
\end{equation*}

Step 4. Screening phase: construct the subset $D$ using the following criterion:
\begin{equation}
	\begin{aligned}
		&D = \{n \in D_{old} \;\\ 
		&\text{and} \; \overline{P_n(k)} \geq \overline{P_l(k)} - G_{nl}(k), \forall\;n \in D_{old}, n \neq l \},\\
		&\text{where}\; G_{nl}(k) = \max\left\{0, \frac{\delta}{2k}\left(\frac{h^2 S_{nl}^2}{\delta^2} - k\right)\right\}
	\end{aligned}
\end{equation}

Step 5. If $|D|$ = 1, then stop, and the $\theta_n \in D$ is the optimal $\theta$ with probability $1-p$. If not, generate one extra replicate value of the performance measure for each $n \in D$, set $k = k + 1$, $D_{old} = D$, and repeat steps 4 and 5.

In the Step 4 above, $\overline{P_n(k)}$ represents the average value of the performance measure calculated with $k$ replications and generated with the $n_{th}$ hyperparameter set. Note that in Step 3, $k = R_0$.

With the KN method, the number of evaluations of the ML model are relatively larger than those of the grid search or the random search method, as the number of `replicate' values of the performance measure to be generated are not \textit{a priori} known. So, the KN method is ideally suited for cases where the hyperparameter space is relatively small, and where generating replicate estimates of the performance measure is not expensive. We demonstrate such use cases in Section \ref{kneval}.

\subsection{Stochastic Ruler Random Search Method Overview}
\label{srover}
The KN method requires complete enumeration and evaluation of the search space. This can become a computationally expensive exercise when the search space cardinality is relatively high, or when generation of replicate values of the objective function via the simulation is expensive. In such situations, random search methods are useful. The stochastic ruler random search method, one of the earliest and simplest random search methods, was developed by\citet{yan1992stochastic} to solve discrete simulation optimisation problems similar to that expressed in equation~\ref{eqsimopt}.

The method involves comparing each replicate objective function value \(h(x)\), where \(x \text{ belongs to the solution space } \mathbb{S}\), with a uniform random variable defined on the possible range of values within which \(h(x)\) lies. This uniform random variable is used as a scale - that is, the stochastic ruler - against which the observations \(h(x)\) are compared. The method converges asymptotically (in probability) to a global optimal solution. In this paper, in addition to the original version of the stochastic ruler method \citep{yan1992stochastic}, we also implement its modification proposed by \citet{ramsr2020}. 

%We begin by providing an overview of the stochastic ruler method.  

The method is initialized with a feasible solution, and then proceeds by searching the neighborhood of the solution for a more suitable solution. A neighbourhood structure is constructed for all $x \in \mathbb{S}$ satisfying notions of `reachability' \citep{yan1992stochastic} such that each solution in the feasible space can be reached from another solution $x'$ which may or may not be in the neighborhood $N(x)$ of $x$. Further, the neighborhood structure must also satisfy a condition of symmetry: if $x' \in N(x)$ then $x \in N(x')$. From a given estimate of the solution $x$, the next estimate is identified by checking whether every replicate value of $h(x')$ ($x' \in N(x)$) generated by the simulation is greater than (for a maximization problem) a sample from a uniform random variable $U(a,b)$ ($a < b$), where $(a, b)$ typically encompasses the range of possible values of $h(x)$. Hence $U(a, b)$ becomes the stochastic ruler. Such a comparison is performed a maximum of $M_k$ times for each $x' \in N(x)$. In the $k^{th}$ iteration, an $x'_{k} \in N(x_k)$ is set to be the next estimate $x_{k+1}$ of the optimal solution $x^*$ if every one of the $M_k$ replicate values of the objective function is greater than every one of the corresponding $M_k$ samples of the stochastic ruler $U(a,b)$.

The $x'_k \in N(x_k)$ are selected from $N(x_k)$ with probability $\frac{1}{|N(x_k)|}$. Note that $M_k$ is specified to be an nondecreasing function of $k$. While the algorithm converges in probability to the global optimum, in practice, it is terminated when a preset computational budget is exhausted, or an acceptable increase (for a maximization problem) in the objective function value is attained. The specific steps involved in the algorithm are described below. All assumptions and definitions associated with the method can be found in \citet{yan1992stochastic}.

\textit{Initialization}. Construction of neighborhood structure $N(x)$ for $x \in \mathbb{S}$, specification of nondecreasing $M_k$, $U(a,b)$ and initial solution $x_0$, and setting $k := 0$.

\textit{Step 1}. For $x_k = x$, choose a new candidate solution $z$ from the neighbourhood $N(x)$ with probability $P\{z \; | \; x\} = \frac{1}{|N(x)|}, \; z \in N(x)$.

\textit{Step 2}. For $z$, generate one replicate value $h(z)$. Then generate a single realization $u$ from $U(a, b)$. If $h(z) < u$, then let $x_{k+1} = x_k$ and go to Step 3. If not, generate another replicate value $h(z)$ and another realization $u$ from $U(a, b)$. If $h(z) > u$, then let $x_{k+1} = x_k$ and go to Step 3. Otherwise continue to generate replicates $h(z)$ and $u$ from $U(a,b)$ and conduct the comparisons. If all $M_k$ tests, $h(z) < u$, fail, then accept $z$ as the next candidate solution and set $x_{k+1} = z$.

\textit{Step 3}. Set $k = k + 1$ and go to Step 1.

Note that in addition to applying the original version of the SR method (described above), we also applied a recent modification proposed by \citet{ramsr2020} to the HPO problem. The modification involves requiring that only a fraction $\alpha$ ($0 < \alpha < 1$, with $\alpha$ ideally set to be greater than 0.5) of the $M_k$ tests be successful for picking a candidate solution from the neighborhood of the current solution as the next estimate of the optimal solution, unlike the original version where every one of the $M_k$ tests is required to be successful. The advantage of the modification is that it prevents promising candidate solutions in the neighborhood of the current solution from being rejected as the next estimate of the optimal solution if it fails only a small fraction of the $M_k$ (e.g., one test) tests.

%We now describe the application of the stochastic ruler method to various types of hyperparameter optimization problems.

\section{Computational Experiments}
\label{compres}
We begin by demonstrating the application of the KN R\&S and stochastic ruler random search method to a wide variety of ML problems. Subsequently, we present the benchmarking of these methods, along with the AH LCRS method, against existing widely used HPO methods.
\subsection{KN Method: Demonstration for HPO}
\label{kneval}
%\subsection{KN Method: Implementation for Hyperparameter Optimization}
%\label{prob_stat}
We applied the KN method to optimize the hyperparameters for support vector machines and feedforward neural network ML methods. We discretized the hyperparameter spaces for each of these ML methods so that the feasible hyperparameter space is suitable for application of the KN method. We chose the `breast cancer Wisconsin dataset' from the Scikit-learn repository \citep{scikit-learn} for all computational experiments in this section. The dataset contained 569 samples in total, where each sample consists of 30 features and a binary label. Thus the objective of the ML exercise here is to train and test an ML model on the dataset such that it is able to classify a sample into one of two categories: `malignant' or `benign', and the SO problem in turn involves finding the hyperparameter set that maximizes average classification accuracy. 

We utilized the public version of the Google Colaboratory platform, using CPUs with a 2.2 gigaHertz clock cycle speed and 12 gigabytes of memory. The results of applying the KN method, implemented as described above, are provided below in Table~\ref{tab:knsvminit}. We provide the results in Table~\ref{tab:knsvminit} to illustrate how the KN method can be applied for HPO - that is, to illustrate how the parameters of the KN method can be specified, how the search space is constructed, etc. We do not describe the ML methods in question themselves in detail  - that is, the SVM and neural network models used in this manuscript, as well as their hyperparameters - as we assume sufficient familiarity with these ML methods on part of the reader.

\begin{table}[htbp]
	\centering
	\caption{KN method implementation for optimizing support vector machine and artificial neural network method hyperparameters. \textit{Notes.} num = number.}
	\begin{tabular}{|R{7.1cm}|R{2cm}|L{3cm}|}
		\hline
		Hyperparameters & \multicolumn{1}{R{2cm}|}{KN method parameters} & \multicolumn{1}{R{3cm}|}{Results} \bigstrut\\
		\hline
		\multicolumn{3}{|c|}{Support vector machine} \bigstrut\\
		\hline
		Kernel $\in \{'rbf', 'linear'\}$ & \multicolumn{1}{l|}{$N = 200,$ $n_o= 10$} & \multicolumn{1}{l|}{Optimal accuracy: 0.968} \bigstrut\\
		%\hline
		Gamma $\in \{0.001, 0.01, 0.1, 0.5, 1, 10, 30, 50, 80, 100\}$ & \multicolumn{1}{l|}{$p = 0.05$, $\delta = 10\%$} & \multicolumn{1}{l|}{Num function evaluations: 2001} \bigstrut\\
		%\hline
		C $\in \{0.01, 0.1, 1, 10, 100, 300, 500, 700, 800, 1000\}$ &       & \multicolumn{1}{l|}{Net runtime: 6709 seconds} \bigstrut\\
		%\hline
		&       & \multicolumn{1}{l|}{Total num iterations ($k$): 10} \bigstrut\\
		\hline
		\multicolumn{3}{|C{12.1cm}|}{Artifical neural network} \bigstrut\\
		\hline
		Hidden layer sizes $\in \{3, 5, 8, 12, 15, 20, 25, 30, 50, 80\}$ & \multicolumn{1}{l|}{$k$ = 600, \(n_o = 10\)} & \multicolumn{1}{l|}{Optimal accuracy: 0.934} \bigstrut\\
		%\hline
		Activation function $\in \{relu, logistic, tanh\}$ & \multicolumn{1}{l|}{$p = 0.05, \delta = 10\%$} & \multicolumn{1}{l|}{Num function evaluations: 9871} \bigstrut\\
		%\hline
		Solver $\in \{adam, sgd\}$ &       & \multicolumn{1}{l|}{Net runtime: 3093 seconds} \bigstrut\\
		%\hline
		Learning rate type $\in \{constant, adaptive\}$ &       & \multicolumn{1}{l|}{Total num iterations ($k$): 195} \bigstrut\\
		%\hline
		Learning rate value (discretized) $\in$ $\{0.0005,0.001,0.01,0.05,0.1\}$ &       &  \bigstrut\\
		\hline
	\end{tabular}%
	\label{tab:knsvminit}%
\end{table}%

We can see from Table~\ref{tab:knsvminit} that the search space for the SO algorithm is constructed by taking the Cartesian product of the sets of allowable values for the hyperparameters. For some continuous parameters, such as the `Gamma' parameter for the SVM, the range may theoretically be unbounded in $\mathbb{R}$; in such cases, the set of allowable values must initially be chosen as a reasonable range, and then discretized. Other parameters are inherently discrete-valued, such as the kernel type for the SVM, or the activation function type for the ANN. Further, we specified the value of $\delta$ for the above computational experiments to be $10\%$, implying that we are indifferent to any hyperparameter set that yields a classification accuracy within 10\% of the classification accuracy yielded by the optimal solution. 

A drawback of applying the KN method for this purpose is that it requires complete enumeration of the solution space; however, as we see from Table~\ref{tab:knsvminit}, the method identifies an optimal solution in reasonable runtimes when the cardinality of the search space is in the order of hundreds.

We note here that the KN method parameters $p$ and $\delta$ control the evolution of the search process for the optimal hyperparameter set. In this sense, they are themselves `hyperparameters' of the KN method; however, unlike ML method hyperparameters, these parameters control the point at which the KN method terminates its search, and the solution identified at this termination step remains the optimal solution with a corresponding statistical guarantee. We investigated how the number of iterations (the final value of $k$ in the execution of the KN procedure) and correspondingly the computational runtime required to arrive at the optimal hyperparameter set changes with $\delta$. We would expect that as the value of $\delta$ decreases, the final value of $k$ would increase given that more precision would be required of the KN method to distinguish the best hyperparameter set from the next best set. This is indeed the case, as depicted in Figure~\ref{F1} below. 

\begin{figure}[hbt!]
	\begin{center}
		\centering
		\includegraphics[scale=2.2]{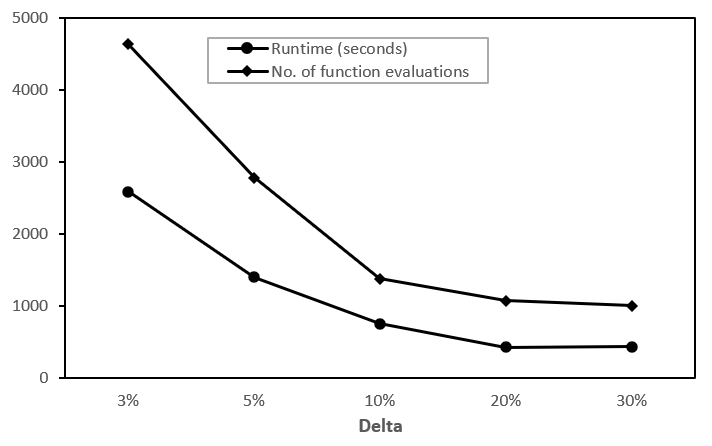}
	\end{center}
	\caption{KN method: number of function evaluations required to arrive at optimal hyperparameter set versus \(\delta\).}
	\label{F1}
\end{figure}

The experiments summarized in Figure~\ref{F1} and in Section~\ref{bnch_KN}, where we describe the process of benchmarking the KN method with respect to HPO methods in the $hyperopt$ package, were conducted with an ANN with the following hyperparameter space. 

\begin{itemize}
	\item Hidden layer sizes $\in \{3, 10, 25, 50, 80\}$
	\item Learning rate value (discretized) $\in$ \{0.0005,0.001,0.01\}
	\item Activation function $\in$ \{relu, logistic, tanh\}
	\item Solver $\in$ \{adam, sgd\}
	\item Learning rate type $\in$ \{adaptive\}
\end{itemize}

\subsection{Stochastic Ruler Method: Demonstration for HPO}
\label{srdemo}
In this section, we demonstrate the application of the stochastic ruler (SR) method to the problem of optimizing the hyperparameters of a variety of neural network machine learning methods and problems: for standard feedforward NNs as described in Section~\ref{kneval}, and for `deep' neural network models such as long short term memory (LSTM) networks for time series prediction and convolutional neural networks (CNNs) for image classification.

%We first demonstrate the application of the SR method to these different types of NN models and problems, and then describe its benchmarking with respect to state-of-the-art hyperparameter optimization methods.

%To test this algorithm, we took a set of machine learning problems which included different types of problems namely classification, image classification and time-series regression. We experimented with different types and size of search space and stopping criterion with MLP. Stages refers to stochastic ruler iterations $k$ in the procedure \ref{procedure}. In these problems, we chose the value of a and b based on our idea of accuracy or loss values specific to the ML models used.

All computational experiments described in this section were executed on the public version of the Google Colaboratory platform, with a 2.2 gigaHertz clock cycle speed CPU and 12 gigabytes of memory.

In the first set of computational experiments, we consider two types of stopping criteria for the SR method. First, we consider an `optimal performance' based criterion: we identify the optimal solution for the HPO problem under consideration on an \textit{a priori} basis by either applying the KN method or running the stochastic ruler method itself for a large duration and terminating the method when it remains in the same neighborhood for a sufficiently long duration. The SR method is then terminated when it reaches this predetermined optimal solution. The second termination criterion we consider resembles a more practical situation when the optimal solution is unknown, and involves terminating the SR method when a preset computational budget is exhausted or after a sufficient improvement in the objective function value is attained. 

Further, we employ two types of neighborhood structures for the solution spaces of the HPO problems considered in this paper. The first involves considering the entire feasible solution space, except for the solution under consideration, to be the neighborhood of each solution. That is, for $x \in \mathbb{S}, N(x) = \mathbb{S} - \{x\}$. We refer to this neighborhood structure as $N_{1(SR)}$. The second neighborhood structure involves considering points in the set of allowable values for each hyperparameter that are adjacent to the hyperparameter value under consideration as neighbors, where the elements of the set are ordered according to some criterion. For example, for the $i^{th}$ hyperparameter taking value $x_{ij}$ in its set of allowable values $\mathbb{S}_i$, its neighborhood is constructed as: $N(x_{ij}) = \{x_{ij} | x_{ij} \in \mathbb{S}_i; j \in \{-1,0,1\}\}$. The neighborhood of a solution $x$ is then constructed as the Cartesian product of the $N(x_i)$ and excluding $x$ itself. That is, $N(x) = \prod\limits_{i = 1}^n N(x_i) - \{x\}$ (where the dimension of the hyperparameter search space is $n$). We refer to this neighborhood structure as $N_{2(SR)}$. Neighborhoods for hyperparameters at the boundary of their ordered set of allowable values are constructed by just taking the corresponding value from the beginning or end of the set. For example, for the $i^{th}$ hyperparameter with set of allowable values $\mathbb{S}_i$, where $|\mathbb{S}_i| = m$, then $N(x_{im}) = \{x_{i,m-1}, x_{im}, x_{i1}\}$.

For all HPO problems discussed in this section, we set $M_k = \frac{\ln{k+10}}{\ln{5}}$.

We begin by demonstrating the implementation of the SR method with the first termination criterion - i.e., the optimal performance criterion. We consider four HPO problems as part of this, and these are listed below.\\

\noindent \textit{Optimal Performance Termination Criterion: Demonstration}.

\begin{enumerate}
	
	\item We use a feedforward ANN with two hidden layers to solve the breast cancer classification problem described in Section \ref{kneval}. We consider the following hyperparameters for the problem, and apply the $N_{2(SR)}$ neighborhood structure for the solution space. We set $U(a, b) = U(0,1)$.\\%We apply the optimal performance stopping criterion here and the results are summarized in table \ref{SR_mlp1_tab}
	
	\noindent Neurons in hidden layer 1 $\in \{2,4,6,8,10\}$\\
	Neurons in hidden layer 2 $\in \{1,2,3,4,5\}$\\
	Learning rate $\in \{0.001, 0.004, 0.007, 0.01\}$\\
	%$U(a,b) = U(0,1)$\\
	
	%Stochastic Ruler procedure parameters:\\
	%\(N(X') = \{X | X \epsilon S; x_i = x'_{i+j}; X \neq X'\}, j \in \{-1,0,1\}\)\\
	%\(a = 0, b = 1\)\\
	
	\item We considered a larger hyperparameter space for the feedforward ANN, considering the activation function and learning rate type hyperparameters as well. We used the $N_{2(SR)}$ neighborhood structure, and set $U(a,b) = U(0,1)$.\\% We use the optimal performance stopping criterion and the results are summarized in table \ref{SR_mlp3_tab}.\\
	
	\noindent Neurons in hidden layer 2 $\in \{2,4,6,8,10\}$\\
	Neurons in hidden layer 1 $\in \{1,2,3,4,5\}$\\
	Learning rate $\in \{0.001, 0.004, 0.007, 0.01\}$\\
	Activation function $\in$ \{relu, tanh, logistic\}\\
	Learning rate type $\in$ \{constant, invscaling, adaptive\}\\
	
	\item We then considered a more complex `deep' neural network model - LSTMs, which are typically used for time series prediction \citep{hochreiter1997long}. For this application of the SR method, we considered a different dataset. This is because each independent variable that forms part of an LSTM feature is a time series in itself. For example, if an LSTM model uses $n$ features, then a single sample in the training dataset for the model would be an $n \times t$ matrix, where each of the $n$ variables is a $t$-dimensional vector. Thus the entire dataset, if it consists of $m$ samples, will be three dimensional. 
	
	For constructing the HPO problem, we applied an LSTM neural network model for stock price prediction. Thus this serves to demonstrate the application of discrete SO methods for regression problems, as opposed to the classification problems that we address in the remainder of this manuscript. We used the Indian National Stock Exchange opening price data for a single stock. We collected the opening stock prices (therefore, a dataset with a single feature) for this stock for a period of 700 days, and divided the data into training and test datasets in a 5:2 ratio. The opening stock price data for a period of 30 consecutive days is used as a single sample used to predict the stock price on the 31\textsuperscript{st} day. Thus a training set consisting of 500 30-day stock price sequences and corresponding regression labels (i.e., the stock prices on the 31\textsuperscript{st} days corresponding to each of the 500 30-day sequences) was used to train the LSTM model. The trained LSTM model's regression accuracy was then estimated using the testing dataset; that is, using 200 30-day sequences and their corresponding labels (the stock price on the 31\textsuperscript{st} day). The hyperparameters for the LSTM model are given below.\\%For testing the results in time-series prediction problems, we applied Stochastic Ruler on LSTM to solve stock price prediction problem as defined below.
	
	%Problem Statement:\\
	%For testing our methods on LSTMs, we used Stock prices of "TATA Global" at NSE, starting from 21st July 2010 to total 700 data points. We used the opening Prices for each day for training and testing the model. We used stock prices of last 30 days to predict the price of the 31st day, sequentially for all the 500 training points, and then tested it on 200 testing points.\\
	
	%Hyperparameters:\\
	\noindent Neurons per layer $\in$ \{10, 50\}\\
	Dropout rate $\in$ \{0.2, 0.5\}\\
	Epochs $\in$ \{5,10,50\}\\
	Batch size $\in$ \{47, 97, 470\}\\
	
	The $N_{2(SR)}$ neighborhood structure along with an $U(0,1)$ stochastic ruler was used. Note that many of the hyperparameters controlling the learning process of LSTM models are different from those of the feedforward ANNs. This is due to the substantial difference in the architecture of LSTMs when compared to feedforward ANNs. We do not describe in detail LSTM model architecture because the focus of this study is demonstrating the application of random search SO methods for deep learning models, and not the deep learning models themselves.
	
	\item We also considered the application of the SR method to optimize the hyperparameters of ANN models commonly used for image processing and computer vision applications - convolutional neural networks (CNNs). CNNs typically take images as input and assign the image into one of two or more predefined categories. We used the MNIST handwritten digit recognition dataset for our demonstration, which is a widely used dataset for benchmarking the performance of CNN models \citep{deng2012mnist}. The classification problem here involves classifying each image of a handwritten digit into one of 10 categories: that is, whether it is one of $\{0,1,2,\dots,9\}$. We used 400 samples from the MNIST dataset available from the Scikit-learn machine learning repository for training the CNN and 100 images for validation. Each sample in the dataset was a $28 \times 28$ pixel image. The $N_{2(SR)}$ neighborhood structure with $U(a,b) = U(0, 0.08)$ was used for this HPO problem. The hyperparameter set for the problem is given below.\\
	
	%We also tested Stochastic Ruler Procedure on CNN model using 'mnist' data set as defined below. We use the optimal performance stopping criterion and the results are summarized in table \ref{SR_cnn_tab}.\\
	
	%Problem Statement:\\
	%For CNN, we used "MNIST" data-set from scikit-learn repository. \\
	%\centering
	%Hyperparameters:\\
	\noindent Neurons per layer $\in$ \{32, 64\}\\
	Epochs $\in$ \{3, 5\}\\
	Batch size $\in$ \{50, 100, 400, 800\}\\  
\end{enumerate}

The results from applying the stochastic ruler method with the optimal performance stopping criterion to the above four neural network HPO problems are summarized in Table~\ref{tab:srnnopt}. Note that `stages' in Table~\ref{tab:srnnopt} and in the remainder of this article refer to the value of $k$ from the description of the SR method - i.e., an evaluation of a specific solution $x \in \mathbb{S}$. The above problems are referred to as Problems 1 through 4 in Table~\ref{tab:srnnopt}.

\begin{table}[htbp]
	\centering
	\caption{Stochastic ruler method for hyperparameter optimization: application to neural networks with optimal performance stopping criterion.}
	\label{tab:srnnopt}
	\begin{tabular}{|c|c|c|}
		\hline
		$\alpha$ & Stages & Runtime (seconds) \bigstrut\\
		\hline
		\multicolumn{3}{|c|}{Problem 1} \bigstrut\\
		\hline
		1     & 1354  & 285 \bigstrut\\
		\hline
		0.8   & 1017  & 212 \bigstrut\\
		\hline
		\multicolumn{3}{|c|}{Problem 2} \bigstrut\\
		\hline
		1     & 3273  & 1463 \bigstrut\\
		\hline
		0.8   & 2712  & 1045 \bigstrut\\
		\hline
		\multicolumn{3}{|c|}{Problem 3} \bigstrut\\
		\hline
		1     & 13    & 88 \bigstrut\\
		\hline
		0.8   & 13    & 83 \bigstrut\\
		\hline
		\multicolumn{3}{|c|}{Problem 4} \bigstrut\\
		\hline
		1     & 10    & 83 \bigstrut\\
		\hline
		0.8   & 9     & 80 \bigstrut\\
		\hline
	\end{tabular}%
\end{table}%

We see from Table~\ref{tab:srnnopt} that in all cases the SR method arrives at the optimal solution within reasonable runtimes. The runtimes for Problem 2 are significantly higher because the number of hyperparameters are also correspondingly higher. We also see that the modified version of the SR method (proposed by \citet{ramsr2020}) performs at least as well as the original method, and outperforms the original method both in terms of runtimes and stages when the dimensionality of the search space is also higher (Problems 1 and 2). 

We now describe the implementation of the SR method for the more realistic HPO case when the optimal solution is not known - that is, when exhaustion of the computational budget is used as the termination criteria. \\

\noindent \textit{Computational Budget Termination Criterion: Demonstration}.\\ \\
We consider two HPO problems (Problems 5 and 6 below) implemented using this termination criterion in this section, and discuss the implementation of this criterion in more detail in the following section where we discuss benchmarking the SR method against state-of-the-art HPO packages. 

The computational budgets for Problems 5 and 6 below are specified in terms of maximum computational runtimes, measured in seconds. We measured the average and standard deviation of the improvement in the classification accuracy when starting with a randomly selected hyperparameter set from the search space. In these problems also, we applied both the original as well as the modified versions of the SR method.

\begin{enumerate}
	\item[5.] We considered the application of the feedforward ANN for the classification problem as described for Problems 1 and 2. We used the $N_{2(SR)}$ neighborhood structure, set $U(a,b) = U(0,1)$, and specified a computational budget of 1,500 seconds. The hyperparameter space is described below.\\% We applied Stochastic Ruler on MLP NN with a larger search space. We noted the mean improvement observed within the computational budget stopping criterion, as summarized in table \ref{SR_mlp2_tab}.\\
	
	%Hyperparameters:\\
	\noindent Neurons in hidden layer 2 $\in \{2,4,6,8,10\}$\\
	Neurons in hidden layer 1 $\in \{1,2,3,4,5\}$\\
	Learning rate $\in \{1e-6, 5e-6, 1e-5, 4e-5, 7e-5, 1e-4, 4e-4, 7e-4, 1e-3\}$\\
	
	\item[6.] We then considered the same HPO problem as in problem 5, but with a discrete-valued search space alone. This is in contrast to previous examples where we discretized the otherwise continuous-valued learning rate search space. We used the $N_{2(SR)}$ neighborhood structure, set $U(a,b) = U(0,1)$ and specified a computational budget of 250 seconds.% Running Stochastic Ruler on MLP NN with only discrete type hyperparameter space. We noted the mean improvement observed within the computational budget stopping criterion, as summarized in table \ref{SR_mlp4_tab}.\\
	
	%Hyperparameters:\\
	\noindent Neurons in hidden layer 2 $\in \{2,4,6,8,10\}$\\
	Neurons in hidden layer 1 $\in \{1,2,3,4,5\}$\\
\end{enumerate}

The results of applying the SR method to Problems 5 and 6 are provided in Table~\ref{tab:srnnbud} below. 
\begin{table}[htbp]
	\centering
	\caption{Stochastic ruler method for hyperparameter optimization: application to neural networks with computational budget based termination criterion.}
	\label{tab:srnnbud}
	\begin{tabular}{|c|c|c|}
		\hline
		$\alpha$ & Stages & \multicolumn{1}{p{13.225em}|}{Mean classification performance improvement (SD)} \bigstrut\\
		\hline
		\multicolumn{3}{|c|}{Problem 1, budget: 1,500 seconds} \bigstrut\\
		\hline
		1     & 2699  & 0.4982 (0.083) \bigstrut\\
		\hline
		0.8   & 2649  & 0.4730 (0.107) \bigstrut\\
		\hline
		0.6   & 1946  & 0.4999 (0.084) \bigstrut\\
		\hline
		\multicolumn{3}{|c|}{Problem 2, budget: 250 seconds} \bigstrut\\
		\hline
		1     & 599   & 0.5747 (0.074) \bigstrut\\
		\hline
		0.8   & 620   & 0.5334 (0.024) \bigstrut\\
		\hline
		0.6   & 585   & 0.5581 (0.065) \bigstrut\\
		\hline
	\end{tabular}%
\end{table}%

We see from Table~\ref{tab:srnnbud} that a significant improvement in classification accuracy is observed on average within the computational budget. We see similar trends regarding the performance of the original and modified versions of the SR algorithm, and for these problems, we see that setting $\alpha = 0.6$ also yields acceptable results. 

%We now discuss the benchmarking of the SR method with existing widely used hyperparameter optimization packages.

\subsection{Benchmarking: the KN Method}
\label{bnch_KN}

We now discuss the benchmarking of the KN method against the RS and the Tree of Parzen Estimators (TPE) methods implemented by the $hyperopt$ library. We have previously provided brief introductions to the $hyperopt$ RS and Bayesian optimization methods in Section~\ref{sec:intro}. The $hyperopt$ TPE method is a commonly used instance of the Bayesian optimization approach for HPO. We remind readers here that the benchmarking exercise for the KN method was carried out with a feedforward ANN with five hyperparameters, and was described in Section~\ref{kneval}. %The $hyperopt$ TPE algorithm \citep{bergstra2013making} is a Bayesian optimization algorithm, which constructs a Gaussian mixture model of the classification performance as a function of hyperparameter sets, and then performs a Bayesian update of the classification performance given a hyperparameter set at each iteration.%

For the KN method, we consider two benchmarking scenarios. In the first scenario, given the fact that the KN method typically finds the optimal solution in finite computational runtimes with a probabilistic guarantee, we let the method run its course to find the optimal solution without imposing a computational budget constraint. On the other hand, for the RS and TPE methods, given that they do not possess this property, they are typically deployed with a computational budget based termination criterion specified in terms of the maximum number of function evaluations. Therefore, this benchmarking exercise was conducted to highlight this difference between these two method types: we compare the performance of the KN method when it is allowed to run to completion with the performance of the RS and TPE methods under multiple computational budget cases. Note that each function evaluation involves generation of one replicate value of the performance measure for a hyperparameter set. Further, for the KN method, multiple function evaluations will be generated for the same hyperparameter set, whereas the $hyperopt$ RS and TPE methods evaluate a hyperparameter set via generation of only one replicate value of the performance measure. %However, the same hyperparameter set may be evaluated more than once if it is randomly sampled more than once. Thus for the RS and TPE methods, we compare the performance of its optimal hyperparameter set for multiple computational budgets, where said budgets are specified in terms of the number of function evaluations.

In the second scenario, we customize the KN method such that it is applied in conjunction with a computational budget specified in terms of number of function evaluations. For instance, if $N$ total function evaluations are available, and the cardinality of the solution space is $K$, then $R_0$ function evaluations are allocated to each solution for the screening phase, where $K \times R_0 < N$. The remainder of the functional evaluations ($N - (K \times R_0)$ evaluations) are reserved for the ranking and selection phase. When the computational budget is exhausted, every solution remaining in the subset $D$ of remaining candidate solutions (see Section~\ref{knover} for the definition of $D$) is equally likely to be the optimal solution. If $|D| > 1$, then we compare the performance of each solution in $D$ with the performance of the optimal solution returned by the RS or TPE methods. From the standpoint of future deployment of this customization of the KN method for HPO in, for example, an HPO library, a subset of the remaining solutions in $D$ may be chosen based on some predefined criterion (e.g., solutions with the top three mean classification accuracies at the time of algorithm termination), and this subset can be provided to the analyst as the top performing hyperparameter sets.

Before we provide the results of the benchmarking exercise for both scenarios above, we note that because the the RS and TPE methods use only a single replicate value of the performance measure to evaluate a hyperparameter set, it is advisable to conduct a more rigorous comparison of the optimal solutions returned by the $hyperopt$ methods and the KN method. Therefore, we conduct a Student's $t$-test, at a 5\% level of significance, to compare the mean performances of the optimal hyperparameter sets as returned by the KN method and the RS and TPE methods, respectively. 25 replicate values of the classification accuracies were generated for the optimal hyperparameter sets identified by the KN method and the sets identified by the RS and TPE methods for each computational budget. $t$-tests for equality of the mean classification accuracies for both hyperparameter sets were conducted using these replications. For the first benchmarking scenario, only a single hyperparameter set is returned as the optimal solution by the KN method, so only a single such test is performed. However, for the second benchmarking scenario, because multiple hyperparameter sets may be returned by the KN method, such tests are conducted for each hyperparameter set with respect to the $hyperopt$ RS or TPE solution.

For the first benchmarking exercise, we compare the classification accuracies of the best system returned by the KN and the $hyperopt$ methods as well as the number of function evaluations required to generate the optimal solution. The results are documented in Table~\ref{KN_benchmark_tab} below. %However, given that the RS and TPE methods use only a single replicate value of the performance measure to evaluate a hyperparameter set, it is advisable to conduct a more rigorous comparison of the optimal solutions returned by the $hyperopt$ methods and the KN method. Therefore, we conduct a Student's $t$-test, at a 5\% level of significance, to compare the mean performances of the optimal hyperparameter sets as returned by the KN method and the RS and TPE methods, respectively. 25 replicate values of the classification accuracies were generated for the optimal hyperparameter set identified by the KN method and the sets identified by the RS and TPE methods for each computational budget. $t$-tests for equality of the mean classification accuracies for both hyperparameter sets were conducted using these replications.

\begin{table}[htbp]
	\centering
	\caption{Benchmarking the KN method versus the $hyperopt$ random search and $hyperopt$ TPE implementation: scenario one (no computational budget for the KN method). \textit{Notes.} RS = random search; TPE = Tree of Parzen Estimators, N/A = not applicable.}
	\label{KN_benchmark_tab} 
	\begin{tabular}{|C{2.3cm}|C{1.5cm}|C{1.5cm}|C{1.5cm}|C{1.8cm}|C{1.5cm}|C{1.8cm}|}
		\hline
		Method & Number of function evaluations & Runtime (seconds) & Optimal objective function value & Mean accuracy from optimal hyperparameter set & $p$ value from $t$-test for equality of means & Inference  \bigstrut\\
		\hline
		KN    & 1381  & 635   & 0.940  & 0.932 & N/A   & N/A  \bigstrut\\
		\hline
		\multirow{5}[10]{*}{$hyperopt$ RS} & 1000  & 463   & 0.982 & 0.815 & 0.001 & KN is better  \bigstrut\\
		\cline{2-7}          & 500   & 209   & 0.982 & 0.89  & 0.018 & KN is better  \bigstrut\\
		\cline{2-7}          & 300   & 124   & 0.982 & 0.728 & 0.000     & KN is better  \bigstrut\\
		\cline{2-7}          & 200   & 85    & 0.972 & 0.655 & 0.000     & KN is better  \bigstrut\\
		\cline{2-7}          & 100   & 42    & 0.965 & 0.941 & 0.187 & KN is comparable \bigstrut\\
		\hline
		\multirow{5}[10]{*}{$hyperopt$ TPE} & 1000  & 463   & 0.991 & 0.815 & 0.001 & KN is better  \bigstrut\\
		\cline{2-7}          & 500   & 306   & 0.991 & 0.927 & 0.724 & KN is comparable \bigstrut\\
		\cline{2-7}          & 300   & 248   & 0.982 & 0.911 & 0.044 & KN is better  \bigstrut\\
		\cline{2-7}          & 200   & 138   & 0.982 & 0.85  & 0.069 & KN is comparable \bigstrut\\
		\cline{2-7}          & 100   & 88    & 0.973 & 0.882 & 0.001 & KN is better  \bigstrut\\
		\hline
	\end{tabular}%%
\end{table}%

We first provide a brief note on the organization of the results in Table~\ref{KN_benchmark_tab}. In the `KN' row (row 2), we provide the results from the implementation of the KN method when it is run to completion. In the next two sets of rows, the results from the implementations of $hyperopt$ RS (rows 3 - 7) and TPE (rows 8 - 12) for various computational budgets (in terms of the maximum number of function evaluations, column 2 from the left) are provided. In column 4 from the left, the optimal objective function value - i.e., classification accuracy - corresponding to the optimal hyperparameter set returned by the HPO method in question is provided. For the $hyperopt$ RS and TPE methods, these correspond to a single replicate measurement of classification accuracy associated with the `optimal' hyperparameter set they return. In column 5, the average classification accuracy calculated from the 25 replicate values of classification accuracy obtained from the ML method with the optimal hyperparameter set (generated for the $t$-test described above) is provided. In column 6, the $p$-value associated with the $t$-test is provided, and in column 7, the inference from the test is provided. With regard to column 7, the term `comparable' indicates that the null hypothesis associated with the $t$-test is not rejected (as evident from the $p$-value in column 6), and the terms `better' or `worse' indicate that the null hypothesis is rejected, and the method with the higher or lower average accuracy is indicated (for example, `KN better' or `KN worse'). Note that this terminology for the statistical comparison between the KN method and the benchmarking baselines is retained for the remainder of the benchmarking experiments presented in this study.

It is evident from Table~\ref{KN_benchmark_tab} that, at minimum, the KN method performs at least as well as all the cases of $hyperopt$ RS and TPE methods when sufficient computational budget is available to let the KN method run to completion. More specifically, the KN method appears to perform significantly better than the $hyperopt$ RS method in this scenario, as it yields statistically better performance than the RS method in almost all comparisons. With respect to the $hyperopt$ TPE method, the KN method offers statistically better performance in three out of five comparisons, and is comparable in the remaining two cases. Importantly, we see that for the comparisons with both methods, the KN method is found to perform better even for the highest computational budget specified for the $hyperopt$ methods. It can also be seen the computational runtime for the KN method, while higher than the corresponding runtimes for the $hyperopt$ methods, is still reasonable, and is on the same order as that for the $hyperopt$ methods with the highest computational budgets. Finally, and perhaps most importantly, the benefit of explicitly considering stochasticity in the ML training and validation process in performing HPO is evident from columns 4 and 5. We see that while the RS and TPE methods identify hyperparameter sets that ostensibly perform very well (for example, accuracies of approximately 99\% in column 4), they are identified on the basis of a single replicate observation of the ML method performance, and hence these high accuracies are not continued to be observed when more replicate measurements of ML method accuracy are used to obtain an average accuracy associated with the hyperparameter set in question. Hence, we observe the following: (a) that the average classification accuracies associated with these `optimal' solutions returned by the TPE and RS methods decrease substantially when more replicate measurements are obtained and averaged, and (b) cases wherein the `optimal' hyperparameter set returned by, for example, the TPE method with a computational budget of 500 evaluations outperforms the hyperparameter set returned by the same method with a budget of 1000 evaluations. Such behavior is not observed for the KN method because it selects solutions via multiple replicate measurements of the ML method performance associated with each hyperparameter set.

The results from the second benchmarking scenario are presented in Table~\ref{knbudget} below. For each computational budget case below, the value of 10 function evaluations (i.e., 10 replicate ML method performance measures) were allocated to each hyperparameter set in the solution space, and all other method settings remained the same as in the first benchmarking scenario. Further, in each case below, we report the proportion of the solutions in the set $D$ returned by the KN method upon termination of the method that performed better, comparably, or worse than the corresponding optimal solutions returned by the $hyperopt$ RS and TPE methods.

\begin{table}[htbp]
	\centering
	\caption{KN method benchmarking: scenario two (with a computational budget for the KN method). \textit{Notes.} Computational budget is in terms of number of function evaluations. RS = random search; TPE = Tree of Parzen Estimators; b/w = between.}
	\begin{tabular}{|C{1.5cm}|C{1.25cm}|C{1.25cm}|C{1.25cm}|C{1.25cm}|C{1.25cm}|C{1.25cm}|}
		\hline
		\multicolumn{1}{|C{1.5cm}|}{\multirow{2}[4]{*}{Budget}} & \multicolumn{3}{C{5cm}|}{KN vs hyperopt RS: $t$-tests b/w each shortlisted KN solution and RS solution} & \multicolumn{3}{C{5cm}|}{KN vs hyperopt TPE: t-tests b/w each shortlisted KN solution and TPE solution} \bigstrut\\
		\cline{2-7}          & \multicolumn{1}{C{1.5cm}|}{KN better (\%)} & \multicolumn{1}{C{1.35cm}|}{KN comparable (\%)} & \multicolumn{1}{C{1.35cm}|}{KN worse (\%)} & \multicolumn{1}{C{1.35cm}|}{KN better (\%)} & \multicolumn{1}{C{1.35cm}|}{KN comparable (\%)} & \multicolumn{1}{C{1.35cm}|}{KN worse (\%)} \bigstrut\\
		\hline
		1000  & 10.4  & 45.4  & 44.2  & 7.7   & 45.2  & 47.1 \bigstrut\\
		\hline
		1100  & 15.6  & 50.4  & 34    & 14.8  & 45.6  & 39.6 \bigstrut\\
		\hline
		1200  & 12.5  & 59.1  & 28.4  & 9.2   & 54.7  & 36.1 \bigstrut\\
		\hline
		1300  & 39.5  & 51.8  & 8.7   & 38.7  & 56    & 5.3 \bigstrut\\
		\hline
	\end{tabular}%
	\label{knbudget}%
\end{table}%

We see from Table~\ref{knbudget} that even in the case with the lowest computational budget (1000 function evaluations), more than 50\% of the solutions performed at least at a statistically comparable level when compared to the solutions from the RS and TPE methods. Further, this proportion increases steadily with an increase in the computational budget, and we see that for the highest computational budget case (selected to be lower than the number of functional evaluations at which the optimal solution is returned by the KN method for benchmarking scenario one), nearly 40\% of the KN method solutions perform better, and more than 90\% perform at least as well when compared to the RS and TPE solutions.

Overall, given that the KN method provides a statistical guarantee of optimality for the solution selected as the best, especially when it is allowed to run to completion, it is evident that ranking and selection methods such as the KN method, where theoretically feasible, can be considered as a serious alternative to the $hyperopt$ RS and TPE methods for HPO. This is particularly the case when the size of the hyperparameter search space is computationally tractable. In this case, the KN method can be allowed to run to completion. Alternatively, if the computational budget is such that the KN method must be deployed as in benchmarking scenario two, then a suitable $R_0$ must be specified to eliminate noncompetitive candidate solutions while leaving enough budget for the ranking and selection phase. The performance of the KN method, as demonstrated in this section, also indicates that more sophisticated state-of-the-art R\&S methods may be considered for HPO, even when solution space cardinalities are significantly larger. For example, certain ranking and selection methods have been used for solution spaces of sizes numbering in the thousands \citep{nelson2010optimization}.

\subsection{Benchmarking: the Stochastic Ruler Method}
\label{bnch_SR}
As discussed in Section~\ref{orgcomp}, we benchmark the SR method against HPO algorithms in the $hyperopt$, $mango$, and $optuna$ packages. Specifically, we benchmark the SR method against the TPE algorithm discussed in Section~\ref{bnch_KN}, the RS algorithm, and the continuous RS algorithm from the $hyperopt$ packages. The $hyperopt$ continuous RS algorithm is the continuous version of the $hyperopt$ RS method \citep{bergstra2012random}, and does not discretize the search space unlike its discrete counterpart. As discussed in the literature review section, $mango$ is a recently developed HPO package that implements a state-of-the-art batch Gaussian process bandit search algorithm \citep{sandha2020mango}. Finally, as discussed in Section~\ref{orgcomp}, we also consider the nondominated sorting genetic algorithm implementation in the $optuna$ package.%The $hyperopt$ TPE algorithm \citep{bergstra2013making} is a Bayesian optimization algorithm, which constructs a Gaussian mixture model of the classification performance as a function of hyperparameter sets, and then performs a Bayesian update of the classification performance given a hyperparameter set at each iteration.

%The algorithm implements an adaptive exploitation vs. exploitation trade-off search strategy, parameterized by the size of the solution space and parallel batch size, among others.%Given that we have already demonstrated the implementation of this termination criterion from the runtime standpoint in the previous section, we choose the latter approach for specifying the computational budget in this benchmarking exercise.

The benchmarking process that we implement for the SR method is similar to that followed for the KN method in the second benchmarking scenario. The SR method only promises asymptotic convergence (in probability) to the global optimal solution, and hence in practical HPO situations, it must be implemented with a computational budget specified in terms of runtime or function evaluations. Thus we specify computational budgets in terms of 100, 200 and 300 function evaluations and compare the performance of the SR method to that of the $hyperopt$, $mango$ and $optuna$ methods for each of these cases. All benchmarking experiments were carried out taking Problem 5 described in Section~\ref{srdemo} as the test case. Secondly, we use two parameterizations of $U(a,b)$: $U(0.5, 1)$ and $U(0.7, 1)$, and provide benchmarking results for both cases. In each case, the SR method was parameterized with a randomly chosen initial solution.%However, for the KN method, we allow it to run its course given that the solution space of the HPO problem that we explored in its benchmarking process was tractable given the computational infrastructure at hand. 

After identifying the hyperparameter set that each method (SR or the benchmark method) converges to within the computational budget, we conduct Student's $t$-tests to check for statistically significant differences between the means of the classification accuracies from both methods. We conduct 10 such tests (by changing random number seeds appropriately, and generating a different set of training and testing dataset permutations corresponding to each random number seed) for each comparison. We then report the number of times the null hypothesis (both means are equal) was rejected, and in such cases, which method's mean was higher. We do not conduct such a rigorous evaluation for the benchmarking of the KN method because of the nature of the method - if it runs to completion (i.e., the set of solutions is iteratively reduced to a singleton set with the optimal solution), the optimal solution is identified with a built-in statistical guarantee. Further, even when the KN method is deployed as in the second benchmarking scenario, a significantly larger number of replications per candidate solution are likely to be used than for the SR method.

The results from the benchmarking experiments are provided in Table~\ref{tab:srbenchmark} below.

% Table generated by Excel2LaTeX from sheet 'Sheet2'
\begin{table}[htbp]
	\centering
	\caption{Benchmarking of the stochastic ruler method against $hyperopt$, $mango$, and $optuna$ HPO methods: first case. \textit{Notes.} Max. = maximum; SR = stochastic ruler; TPE = tree of Parzen estimators; RS = random search; GA = genetic algorithm; N/A = not applicable.}
	\label{tab:srbenchmark}
	\begin{tabular}{|c|c|c|c|}
		\hline
		\multicolumn{1}{|C{2.5cm}|}{Stochastic ruler versus:} & \multicolumn{1}{p{6.09em}|}{Max. function  evaluations} & \multicolumn{1}{p{8.5em}|}{Proportion of tests with p $<$ 0.05} & \multicolumn{1}{p{10.865em}|}{If null rejected, which model has higher mean performance?} \bigstrut\\
		\hline
		\multicolumn{4}{|c|}{$U(a,b) = U(0.7, 1.0)$} \bigstrut\\
		\hline
		\multirow{3}[6]{*}{$hyperopt$ TPE} & 100   & 0.3   & $hyperopt$ TPE \bigstrut\\
		\cline{2-4}          & 200   & 0.0   & N/A \bigstrut\\
		\cline{2-4}          & 300   & 0.0   & N/A \bigstrut\\
		\hline
		\multirow{2}[4]{*}{$hyperopt$ RS} & 100   & 0.7   & SR \bigstrut\\
		\cline{2-4}          & 200   & 1.0   & SR \bigstrut\\
		\hline
		\multirow{2}[4]{*}{$hyperopt$ continuous RS} & 100   & 0.2   & $hyperopt$ continuous RS \bigstrut\\
		\cline{2-4}          & 200   & 0.0   & N/A \bigstrut\\
		\hline
		$mango$ & 200   & 0.0   & N/A \bigstrut\\
		\hline
		\multirow{5}[10]{*}{$optuna$ GA} & 100   & 1     & $optuna$ 90\% times \bigstrut\\
		\cline{2-4}          & 200   & 0.7   & $optuna$ GA 86\% \bigstrut\\
		\cline{2-4}          & 300   & 0.4   & $optuna$ \bigstrut\\
		\cline{2-4}          & 400   & 0.6   & SR 66\% \bigstrut\\
		\cline{2-4}          & 500   & 0.4   & $optuna$ GA \bigstrut\\
		\hline
		\multicolumn{4}{|c|}{$U(a,b) = U(0.5, 1.0)$} \bigstrut\\
		\hline
		\multirow{3}[6]{*}{$hyperopt$ TPE} & 100   & 0.4   & $hyperopt$ TPE \bigstrut\\
		\cline{2-4}          & 200   & 0.2   & $hyperopt$ TPE \bigstrut\\
		\cline{2-4}          & 300   & 0.0   & N/A \bigstrut\\
		\hline
		\multirow{2}[4]{*}{$hyperopt$ RS} & 100   & 0.6   & SR \bigstrut\\
		\cline{2-4}          & 200   & 0.8   & SR \bigstrut\\
		\hline
		\multirow{2}[4]{*}{$hyperopt$ continuous RS} & 100   & 0.4   & $hyperopt$ continuous RS \bigstrut\\
		\cline{2-4}          & 200   & 0.3   & SR \bigstrut\\
		\hline
		\multirow{2}[4]{*}{$mango$} & 200   & 0.2   & $mango$ \bigstrut\\
		\cline{2-4}          & 300   & 0.0   & N/A \bigstrut\\
		\hline
		\multirow{5}[10]{*}{$optuna$ GA} & 100   & 0.4   & $optuna$ GA \bigstrut\\
		\cline{2-4}          & 200   & 0.4   & $optuna$ GA \bigstrut\\
		\cline{2-4}          & 300   & 0.4   & 50\% SR \bigstrut\\
		\cline{2-4}          & 400   & 0.3   & 66\% $optuna$ GA \bigstrut\\
		\cline{2-4}          & 500   & 0.3   & $optuna$ GA \bigstrut\\
		\hline
	\end{tabular}%%
\end{table}%

We observe from Table~\ref{tab:srbenchmark} that the SR method offers comparable performance with respect to the $hyperopt$ and $mango$ HPO methods. In particular, we consistently observe that with higher computational budgets (200 function evaluations and above), the SR method identifies optimal hyperparameter sets that offer classification accuracies that are statistically comparable to every $hyperopt$ and $mango$ method evaluated. The advantage that SO methods offer over $hyperopt$'s RS method is clear. We also see that even for smaller computational budgets (100 function evaluations), the proportion of tests where the $hyperopt$ or $mango$ methods appear to perform better is less than 0.5. Finally, we note that a more `informatively' parameterized stochastic ruler - that is, $U(0.7,1.0)$, which reflects the information that the classification performance with most hyperparameter sets is unlikely to fall below 70\% - appears to lead to better performance for the SR method with respect to the $hyperopt$ and $mango$ methods. With respect to the $optuna$ GA method, we see that with the more informative stochastic ruler - that is, with $U(0.7,1.0)$ - the $optuna$ GA method outperforms the SR method at budgets up to 200 function evaluations. For higher computational budgets, the performance of the SR method appears comparable to that of the $optuna$ GA method. Interestingly, with the `less informative' stochastic ruler ($U(0.5,1.0)$), we see that the performance of the SR method is better than in the previous case, and is largely equivalent to that of the $optuna$ GA method.

In addition to the above benchmarking exercise for the SR method, we also consider a HPO setting with a larger hyperparameter space. For this benchmarking case, we again consider a feedforward ANN with the following hyperparameter space. \\

\noindent Number of hidden layers $\in \{1,2\}$\\
Neurons in hidden layers $\in \{3, 5, 8, 12, 15, 20, 25, 30, 50, 80\}$\\
Solver $\in$ \{adam, sgd\}\\
Learning rate $\in \{0.0005, 0.001, 0.01, 0.05, 0.10\}$\\
Activation function $\in$ \{relu, tanh, logistic\}\\
Learning rate type $\in$ \{constant, adaptive\}\\

This yields a hyperparameter space of cardinality 6600, as 110 different hidden layer architectures are possible. We omit the $hyperopt$ RS methods from the benchmarking process for this case given the SR method's clear superiority as evidenced from the previous case. The benchmarking process for this case is the same as for the first case, except for the fact that we use larger computational budgets given the larger size of the hyperparameter space. The results are provided in Table~\ref{tab:srbenchmark2} below.

% Table generated by Excel2LaTeX from sheet 'Sheet3'
\begin{table}[htbp]
	\centering
	\caption{Benchmarking of the stochastic ruler method against $hyperopt$, $mango$, and $optuna$ HPO methods: second case. \textit{Notes.} Max. = maximum; SR = stochastic ruler; TPE = tree of Parzen estimators; GA = genetic algorithm.}
	\begin{tabular}{|c|c|c|c|}
		\hline
		\multicolumn{1}{|C{2.5cm}|}{Stochastic ruler versus:} & \multicolumn{1}{p{6.775em}|}{Max. function  evaluations} & \multicolumn{1}{p{7.41em}|}{Proportion of tests with p $<$ 0.05} & \multicolumn{1}{p{12.045em}|}{If null rejected, which model has higher mean performance?} \bigstrut\\
		\hline
		\multicolumn{4}{|c|}{$U(a,b) = U(0.7, 1)$} \bigstrut\\
		\hline
		\multirow{3}[6]{*}{$hyperopt$ TPE} & 400   & 0.3   & 66\% SR \bigstrut\\
		\cline{2-4}          & 800   & 0.6   & 66\% $hyperopt$ TPE \bigstrut\\
		\cline{2-4}          & 1200  & 0.5   & 80\% $hyperopt$ TPE \bigstrut\\
		\hline
		\multirow{3}[6]{*}{$mango$} & 400   & 0.5   & $mango$ \bigstrut\\
		\cline{2-4}          & 800   & 0.5   & 80\% $mango$ \bigstrut\\
		\cline{2-4}          & 1200  & 0.2   & 50\% SR \bigstrut\\
		\hline
		\multirow{3}[6]{*}{$optuna$ GA} & 400   & 0.4   & 50\% SR \bigstrut\\
		\cline{2-4}          & 800   & 0.3   & $optuna$ GA \bigstrut\\
		\cline{2-4}          & 1200  & 0.4   & $optuna$ GA \bigstrut\\
		\hline
		\multicolumn{4}{|c|}{$U(a,b) = U(0.5, 1)$} \bigstrut\\
		\hline
		\multirow{3}[6]{*}{$hyperopt$ TPE} & 400   & 0.5   & 80\% $hyperopt$ TPE \bigstrut\\
		\cline{2-4}          & 800   & 0.5   & 60\% TPE \bigstrut\\
		\cline{2-4}          & 1200  & 0.5   & TPE \bigstrut\\
		\hline
		\multirow{3}[6]{*}{$mango$} & 400   & 0.8   & 87.5\% $mango$ \bigstrut\\
		\cline{2-4}          & 800   & 0.4   & 75\% $mango$ \bigstrut\\
		\cline{2-4}          & 1200  & 0.4   & 50\% SR \bigstrut\\
		\hline
		\multirow{3}[6]{*}{$optuna$ GA} & 400   & 0.4   & 75\% $optuna$ GA \bigstrut\\
		\cline{2-4}          & 800   & 0.4   & 75\% $optuna$ GA \bigstrut\\
		\cline{2-4}          & 1200  & 0.4   & 75\% $optuna$ GA \bigstrut\\
		\hline
	\end{tabular}%
	\label{tab:srbenchmark2}%
\end{table}%

For this hyperparameter space, we see that the SR method performs in a largely comparable manner to the $hyperopt$ TPE method. For both stochastic ruler cases, we see that the $hyperopt$ TPE method outperforms the SR method in a maximum of 50\% of the evaluations. With respect to the $mango$ multi-fidelity method, we see that the SR method performs better than with respect to $hyperopt$ TPE. This is evidenced by the fact that even for the highest computational budget case (1200 evaluations), we see that the SR method performs at least as well as the $mango$ method in up to 90\% of the cases (across both stochastic ruler parameterizations). With respect to the $optuna$ GA method, we see that the performance of the SR method is reasonable, with the SR method being outperformed in a maximum of 40\% of the evaluations.

Overall, we see that the SR method consistently outperforms the $hyperopt$ RS, TPE, and the $mango$ methods for hyperparameter spaces that are, relatively speaking (e.g., an average of approximately four hyperparameters, per the numerical experimentation in \cite{yang2020hyperparameter}), not large. For larger and higher-dimensional hyperparameter spaces, as demonstrated in the second benchmarking case above, we see that the SR method's performance is not as consistent. Further, the $optuna$ GA method appears to marginally outperform the SR method in both benchmarking cases above. This motivates the consideration of random search SO methods designed specifically for such hyperparameter spaces. In this context, in Appendix~\ref{ap:aha}, we discuss the benchmarking of the AH method for the HPO problem considered in the second benchmarking case for the SR method.%~\ref{ap:aha}

%We now discuss the application of random search simulation optimization methods - specifically the stochastic ruler method - to the hyperparameter optimization problem.

%\section{Numerical Evaluation: Stochastic Ruler Random Search Method}
%\label{srnumres}

%\subsection{Computational Experiments}
%\label{srcomp}
\section{Concluding Remarks}
\label{conc}
In this work, we demonstrate the use of simulation optimization algorithms - in particular, those developed or commonly used for discrete search spaces - for selecting hyperparameter sets that maximize the performance of machine learning methods. We provide a comprehensive demonstration of the use of two discrete simulation methods in particular: (a) the KN ranking and selection method, which considers every single solution in the discrete solution space, and provides a statistical guarantee of selecting the optimal hyperparameter set, and (b) the stochastic ruler random search method that moves between neighborhoods of promising candidate solutions. We demonstrate the use of these methods across a wide variety of machine learning methods, datasets and their associated hyperparameter sets - we consider both standard machine learning methods such as support vector machines as well neural network based `deep' learning machine learning models. 

We also provide a reasonably comprehensive benchmarking of these methods against the major classes of existing HPO methods implemented in widely used hyperparameter tuning libraries. In particular, as part of the benchmarking process of the KN method, we also develop and demonstrate a customization of the KN method that involves imposing a computational budget on the method leading to its termination before it runs to completion. This customization may prove useful in other discrete SO settings as well. Further, we also benchmark the AH method against state-of-the-art HPO methods for the case of a large hyperparameter space. After presenting the results from each benchmarking exercise for each method, we also discuss the performance of the SO methods in the benchmarking exercise, and whether they have the potential to provide competitive alternatives to the extant HPO methods as implemented in their Python libraries. Overall, it appears that SO methods have the potential to provide highly competitive alternatives to extant HPO methods and libraries, especially in light of the fact that our implementations are not professionally optimized - from a software development standpoint - versions of these methods.

%Through our computational experiments, we provide evidence that the KN and stochastic ruler methods perform at least as well as state-of-the-art hyperparameter tuning packages such as $hyperopt$, and $mango$, \textcolor{red}{especially for hyperparameter search spaces that are of 3 to 4 dimensions}. More specifically, \textcolor{red}{for such search spaces,} we show consistently better performance when compared to the $hyperopt$'s implementation of both discrete as well as continuous random search \citep{bergstra2012random} hyperparameter tuning methods. With the KN method, we observe statistical evidence of superior performance with respect to more sophisticated methods such as $hyperopt$'s Tree of Parzen Estimators algorithm, and with the stochastic ruler method, we observe statistically comparable performance. We observe statistically comparable performance for the stochastic ruler method with respect to the relatively recently developed $mango$ Gaussian process bandit search method. We note that such performance is observed even though our implementations of the KN and stochastic ruler methods are not professionally optimized - from a software development standpoint - versions of these methods.

Our work has a few limitations, which also serve to suggest avenues for future research. As a first step towards exploring the use of SO methods for HPO problems, we have only explored the use of methods primarily used for discrete SO problems. Thus exploring whether SO methods developed for continuous or mixed-integer solution spaces can be adapted for HPO problems may serve as a future avenue of research. Similarly, we have also only explored the use of a relatively simple R\&S method in this work: the KN method. Given its success for hyperparameter search spaces that are not too large, exploring whether more advanced versions of the KN method or R\&S methods in general could prove a fruitful avenue of future research for HPO. For example, efficient ranking and selection methods for computing in parallel environments have been developed \citep{ni2017efficient}, and exploring the adaptation of these for HPO problems may prove useful.

The dimensionality of the hyperparameter search spaces that we have considered has also been small to moderate from the perspective of simulation optimization. For example, the largest space we have considered is of six dimensions, and for this search space, we find that the stochastic ruler method and its modification required approximately 15-25 minutes of computational runtime to arrive at the optimal hyperparameter set. For most machine learning exercises, the number of hyperparameters that analysts attempt to tune may be of this order, given that only a few hyperparameters usually have a significant effect on the performance of the ML method. This is consistent with both the discussion and numerical experimentation in the review by \cite{yang2020hyperparameter}, where the authors on average considered approximately four hyperparameters in total. For larger hyperparameter search spaces, as demonstrated in Appendix~\ref{ap:aha}, the AH method provides suitable recourse.%~\ref{ap:aha}

Finally, while we have demonstrated the use of SO methods for optimizing the hyperparameters of a reasonably wide variety of machine learning methods, ranging from support vector machines to `deep' neural network models such as convolutional neural networks and long short term memory networks, the use of such methods to even more complex deep learning models such as generative adversarial neural networks remains to be explored, and can offer another avenue of future research.

\section*{Declaration of Competing Interests}
The authors declare that they have no relevant financial or non-financial competing interests to report with regard to this work.

\bibliographystyle{apacite}
\bibliography{sample-base}
%\begin{comment}

\newpage
\appendix
\section{Benchmarking: the Adaptive Hyperbox Method}
\label{ap:aha}
\setcounter{page}{1}
In this section, we first provide an overview of the adaptive hyperbox (AH) method, which is an example of a locally convergent random search (LCRS) random search method \citep{xu2013adaptive}. We then describe the benchmarking of the AH method against the second benchmarking case used for the stochastic ruler (SR) method, as described in Section~\ref{bnch_SR}. This benchmarking case involves six hyperparameters and a hyperparameter search space with cardinality 6,600. Hence we consider an SO method more suited to such larger hyperparameter spaces: the AH method, which has been demonstrated to outperform the COMPASS method \citep{hong2006discrete} for SO problems with more than 10 decision variables, and to provide high-quality solutions in reasonable runtimes for SO problems with up to 100 variables).%As discussed in Section~\ref{bnch_SR}, the performance of the SR method is not as consistent with respect to the baseline HPO methods for this benchmarking case as compared to the first benchmarking case with a smaller hyperparameter space.

We now describe the essentials of the AH method, and do not provide comprehensive mathematical detail regarding the method to contain the length of the manuscript. The AH method is an LCRS method, meaning that used on its own, it guarantees convergence to locally optimum solutions for discrete SO problems. A local optimum in this discrete SO setting is defined as the solution that performs best among all neighbors of the current estimate of the solution, where the neighborhood of a solution is defined as the set of all of its adjacent solutions. 

Locally convergent SO methods typically have three stages: (a) a global search phase that identifies promising regions of the search space; (b) a local search phase that attempts to find local optima from the regions identified in the global phase; and (c) a clean-up phase which typically involves applying an R\&S method to identify the best solution among the local optima estimates identified as part of the local phase. SO methods such as COMPASS, industrial strength COMPASS and the AH algorithm are all LCRS methods that perform the local search phase of the three-phase framework described above. The AH algorithm was developed to provide a better performing local search phase alternative to the COMPASS algorithm for higher-dimensional search spaces.

The AH method has the following structure, adapted from \cite{xu2013adaptive}.

\begin{itemize}
	\item Step 0. Initialize with a randomly chosen solution $x_0$ from a given search space subset identified in the global search phase. Let the iteration count $k$ be set to 0. Generate $n(x_0)$ simulation observations from $x_0$, calculate the mean $\bar{G}(x_0)$ of these observations, and record the total number of observations obtained from $x_0$ as $N(x_0) = n(x_0)$. Set the best performing solution $x_{opt(0)}$ at the current stage equal to $x_0$. Let the set of all solutions visited until and including iteration $k$, denoted by $B(k)$, and the set of solutions from which observations are generated in iteration $k$, denoted by $B_k$, be set to $\{x_0\}$.
	
	\item  Step 1. Let $k = k + 1$. Construct the most promising area (MPA) at iteration $k$, denoted by $\psi_k$, as the intersection of a hyperbox $H_k$ around $x_{opt(k-1)}$ and the search space $\mathbb{S}$. The reader may refer to \cite{xu2013adaptive} for the details of how the hyperbox $H_k$ is constructed around $x_{opt(k)}$. Sample a set of $m$ distinct solutions $\{x_{k1}, x_{k2},\dots,x_{km}\}$ from $\psi_k$ according to a sampling distribution (the uniform distribution may be used), which determines $B_k$ as follows: $B_k$ = $\{x_{k1}, x_{k2},\dots,x_{km}\} \cup \{x_{opt(k-1)}\}$. Set $B(k) = B(k-1) \cup B_k$.
		
	\item Step 2. For each $x_{ki} \in B_k$, generate $n(x_{ki})$ simulation observations, and update the values of $N(x_{ki})$ and $\bar{G}(x_{ki})$.
	
	\item Step 3. Set $x_{opt(k)} = \text{arg } \underset{x\in B_k}{\min}~ \bar{G}(x)$. Go to Step 1.
 \end{itemize}

From the above structure, it can be seen that the number of solutions $m$ to be sampled from the MPA in each iteration and the number of observations to be generated from a solution $x$ in each iteration $n(x)$ are parameters of the method. In our benchmarking exercise, we set the value of $m$ as three, and $n(x_0)$ is determined by the function $min~\{5, \ceil{5(\log k)^{1.01}}\}$.

In our benchmarking of the AH method against the $hyperopt$ TPE, $mango$ and $optuna$ GA HPO methods, we apply the AH method in its most basic form: that is, only in its local search phase form. That is, we do not include a global search phase or an R\&S clean-up phase in its implementation. Thus the method is initialized with a randomly chosen hyperparameter set from the solution space, and is terminated after a computational budget specified in terms of the number of function evaluations is reached. The $x_{opt(k)}$ at this point is returned as the optimal hyperparameter set. Finally, as mentioned before, we consider the same HPO problem as done in the second benchmarking case for the SR method. The results from this benchmarking exercise are summarized in Table \ref{tab:ahabench} below.

% Table generated by Excel2LaTeX from sheet 'Sheet3'
\begin{table}[htbp]
	\centering
	\caption{AH method benchmarking results. \textit{Notes.} AHA = adaptive hyperbox algorithm; Max. = maximum; TPE = tree of Parzen estimators; GA = genetic algorithm; N/A = not applicable. }
	\begin{tabular}{|c|c|c|c|}
		\hline
		\multicolumn{1}{|p{6.82em}|}{AHA versus:} & \multicolumn{1}{p{6.545em}|}{Max. function  evaluations} & \multicolumn{1}{p{7.045em}|}{Proportion of tests with $p$ $<$ 0.05} & \multicolumn{1}{p{10.455em}|}{If null rejected, which model has higher mean performance?} \bigstrut\\
		\hline
		\multirow{3}[6]{*}{$hyperopt$ TPE} & 200   & 0.6   & 50\% AHA \bigstrut\\
		\cline{2-4}          & 400   & 0.7   & 65\% AHA \bigstrut\\
		\cline{2-4}          & 800   & 0.3   & 66\% AHA \bigstrut\\
		\hline
		\multirow{3}[6]{*}{$mango$} & 200   & 0.3   & $mango$ \bigstrut\\
		\cline{2-4}          & 400   & 0.3   & 66\% AHA \bigstrut\\
		\cline{2-4}          & 800   & 0.5   & 60\% $mango$ \bigstrut\\
		\hline
		\multirow{3}[6]{*}{$optuna$ GA} & 200   & 0.2   & 50\% AHA \bigstrut\\
		\cline{2-4}          & 400   & 0.4   & 50\% AHA \bigstrut\\
		\cline{2-4}          & 800   & 0.3   & $optuna$ GA \bigstrut\\
		\hline
	\end{tabular}%
	\label{tab:ahabench}%
\end{table}%

We immediately see from Table~\ref{tab:ahabench} that the performance of the AH method is significantly better with respect to the baseline HPO methods when compared to the performance of the SR method. For example, it is clear that the AH method outperforms the $hyperopt$ TPE method and is comparable to the $mango$ method. Even at the higher computational budget case, we see that the AH method is at least comparable to the $mango$ method in 7 out of 10 cases, compared to 8 out of 10 for the $mango$ method. The AH method's performance clearly outpeforms the $optuna$ GA method in 2 out of 3 computational budget cases, and in the third case, it performs comparably in 7 out of 10 cases. This result is not unexpected given that we implement the AH method in its basic form, without the global search phase and R\&S method. Hence for higher computational budget cases, if the AH method is implemented in its comprehensive form, it is likely to consistently outperform or at least be comparable to the $optuna$ GA method in almost all cases. Overall, the AH method appears to provide a highly competitive alternative to all baseline HPO methods considered in this analysis, even when hyperparameter search spaces are large.
\end{document}